\documentclass[sigconf]{acmart}
\usepackage{amsthm,amsmath,amssymb}
\usepackage{appendix}
\usepackage{multirow}
\usepackage{enumerate}
\usepackage{breakurl}
\usepackage{mathrsfs}
\usepackage{graphicx}
\usepackage{epstopdf}
\usepackage{subfigure}
\usepackage{stfloats}
\usepackage{url}
\usepackage{balance}

\usepackage{bm}
\usepackage{color}
\AtBeginDocument{%
  \providecommand\BibTeX{{%
    \normalfont B\kern-0.5em{\scshape i\kern-0.25em b}\kern-0.8em\TeX}}}

\copyrightyear{2020}
\acmYear{2020}
\setcopyright{acmcopyright}\acmConference[KDD '20]{Proceedings of the 26th ACM SIGKDD Conference on Knowledge Discovery and Data Mining USB Stick}{August 23--27, 2020}{Virtual Event, USA}
\acmBooktitle{Proceedings of the 26th ACM SIGKDD Conference on Knowledge Discovery and Data Mining USB Stick (KDD '20), August 23--27, 2020, Virtual Event, USA}
\acmPrice{15.00}
\acmDOI{10.1145/3394486.3403177}
\acmISBN{978-1-4503-7998-4/20/08}

\newcommand{\nop}[1]{}

\acmSubmissionID{1297}

\begin{document}
\fancyhead{}
%%
%% The "title" command has an optional parameter,
%% allowing the author to define a "short title" to be used in page headers.
\title{AM-GCN: Adaptive Multi-channel Graph Convolutional Networks}

%%
%% The "author" command and its associated commands are used to define
%% the authors and their affiliations.
%% Of note is the shared affiliation of the first two authors, and the
%% "authornote" and "authornotemark" commands
%% used to denote shared contribution to the research.
\author{Xiao Wang}
\affiliation{Beijing University of Posts and Telecommunications}
\email{xiaowang@bupt.edu.cn}

\author{Meiqi Zhu}
\affiliation{Beijing University of Posts and Telecommunications}
\email{zhumeiqi@bupt.edu.cn}

\author{Deyu Bo}
\affiliation{Beijing University of Posts and Telecommunications}
\email{bodeyu@bupt.edu.cn}

\author{Peng Cui}
\affiliation{Tsinghua University}
\email{cuip@tsinghua.edu.cn}

\author{Chuan Shi}
\authornote{Corresponding author}
\affiliation{Beijing University of Posts and Telecommunications}
\email{shichuan@bupt.edu.cn}

\author{Jian Pei}
\affiliation{Simon Fraser University}
\email{jpei@cs.sfu.ca}
%%
%% By default, the full list of authors will be used in the page
%% headers. Often, this list is too long, and will overlap
%% other information printed in the page headers. This command allows
%% the author to define a more concise list
%% of authors' names for this purpose.
\renewcommand{\shortauthors}{Wang, et al.}

%%
%% The abstract is a short summary of the work to be presented in the
%% article.
\begin{abstract}

Graph Convolutional Networks (GCNs) have gained great popularity in tackling various analytics tasks on graph and network data.  However, some recent studies raise concerns about whether GCNs can optimally integrate node features and topological structures in a complex graph with rich information. In this paper, we first present an experimental investigation.  Surprisingly, our experimental results clearly show that the capability of the state-of-the-art GCNs in fusing node features and topological structures is distant from optimal or even satisfactory. The weakness may severely hinder the capability of GCNs in some classification tasks, since GCNs may not be able to adaptively learn some deep correlation information between topological structures and node features. Can we remedy the weakness and design a new type of GCNs that can retain the advantages of the state-of-the-art GCNs and, at the same time, enhance the capability of fusing topological structures and node features substantially? We tackle the challenge and propose an adaptive multi-channel graph convolutional networks for semi-supervised classification (AM-GCN).  The central idea is that we extract the specific and common embeddings from node features, topological structures, and their combinations simultaneously, and use the attention mechanism to learn adaptive importance weights of the embeddings.  Our extensive experiments on benchmark data sets clearly show that AM-GCN extracts the most correlated information from both node features and topological structures substantially, and improves the classification accuracy with a clear margin.

\end{abstract}

%%
%% The code below is generated by the tool at http://dl.acm.org/ccs.cfm.
%% Please copy and paste the code instead of the example below.
%%



%%
%% This command processes the author and affiliation and title
%% information and builds the first part of the formatted document.
\maketitle

\section{Introduction}

Network data is ubiquitous, such as social networks, biology networks, and citation networks. Recently, Graph Convolutional Networks (GCNs), a class of neural networks designed to learn graph data, have shown great popularity in tackling graph analytics problems, such as node classification~\cite{abu-el-haija2019mixhop, wu2019demo}, graph classification~\cite{gao2019graph, zhang2018an}, link prediction~\cite{you2019position, kipf2016variational} and recommendation~\cite{ying2018graph, fan2019graph}.

The typical GCN~\cite{kipf2017semi} and its variants~\cite{ve2018graph, hamilton2017inductive, ma2019disentangled, you2019position, wu2019simplifying} usually follow a message-passing manner. A key step is feature aggregation, i.e., a node aggregates feature information from its topological neighbors in each convolutional layer. In this way, feature information propagates over network topology to node embedding, and then node embedding learned as such is used in classification tasks. The whole process is supervised partially by the node labels. The enormous success of GCN is partially thanks to that GCN provides a fusion strategy on topological structures and node features to learn node embedding, and the fusion process is supervised by an end-to-end learning framework.

Some recent studies, however, disclose certain weakness of the state-of-the-art GCNs in fusing node features and topological structures. For example, Li~\textit{et al.}~\cite{li2018deeper} show that GCNs actually perform the Laplacian smoothing on node features, and make the node embedding in the whole network gradually converge.  Nt and Maehara~\cite{nt2019revisiting} and Wu~\textit{et al.}~\cite{wu2019simplifying} prove that topological structures play the role of low-pass filtering on node features when the feature information propagates over network topological structure. Gao~\textit{et~al.}~\cite{gao2019conditional} design a Conditional Random Field (CRF) layer in GCN to explicitly preserve connectivity between nodes.

\emph{What information do GCNs really learn and fuse from topological structures and node features?}  This is a fundamental question since GCNs are often used as an end-to-end learning framework.  A well informed answer to this question can help us understand the capability and limitations of GCNs in a principled way.  This motivates our study immediately.
%To some degree, there is very little consensus about what information on earth GCN truly learns from topology and node feature. Therefore, we reasonably doubt whether GCN is able to adaptively fuse topology and feature to learn the most correlated node embedding for the classification task. Many people would consider that because GCN is a typical end-to-end learning framework, the supervision ability of node label will support GCN to automatically extract the effective information from both topology and node feature. However, with the following two simulation experiments, we think that the answer may need second thought.

As the first contribution of this study, we present experiments assessing the capability of GCNs in fusing topological structures and node features.  Surprisingly, our experiments clearly show that the fusion capability of GCNs on network topological structures and node features is clearly distant from optimal or even satisfactory. Even under some simple situations that the correlation between node features/topology with node label is very clear, GCNs still cannot adequately fuse node features and topological structures to extract the most correlated information (shown in Section \ref{sec:capability}). The weakness may severely hinder the capability of GCNs in some classification tasks, since GCNs may not be able to adaptively learn some correlation information between topological structures and node features.

Once the weakness of the state-of-the-art GCNs in fusion is identified, a natural question is, ``\emph{Can we remedy the weakness and design a new type of GCNs that can retain the advantages of the state-of-the-art GCNs and, at the same time, enhance the capability of fusing topological structures and node features substantially?}''

A good fusion capability of GCNs should substantially extract and fuse the most correlated information for classification task, however, one biggest obstacle in reality is that the correlation between network data and classification task is usually very complex and agnostic. The classification can be correlated with either the topology, or node features, or their combinations. This paper tackles the challenge and proposes an adaptive multi-channel graph convolutional networks for semi-supervised classification (AM-GCN).  The central idea is that we learn the node embedding based on node features, topological structures, and their combinations simultaneously.  The rationale is that the similarity between features and that inferred by topological structures are complementary to each other and can be fused adaptively to derive deeper correlation information for classification tasks.

Technically, in order to fully exploit the information in feature space, we derive the \textit{k}-nearest neighbor graph generated from node features as the feature structural graph.
%Since network topology and features may not be completely independent from each other, there may be same and different characters between them. Therefore,
With the feature graph and the topology graph, we propagate node features over both topology space and feature space, so as to extract two specific embeddings in these two spaces with two specific convolution modules. Considering the common characteristics between two spaces, we design a common convolution module with a parameter sharing strategy to extract the common embedding shared by them. We further utilize the attention mechanism to automatically learn the importance weights for different embeddings, so as to adaptively fuse them. In this way, node labels are able to supervise the learning process to adaptively adjust the weight to extract the most correlated information. Moreover, we design the consistency and disparity constraints to ensure the consistency and disparity of the learned embeddings.

We summarize our main contributions as follows:
\begin{itemize}
\vspace{-\topsep}\item We present experiments
assessing the capability of GCNs in fusing topological structures and node features and identify the weakness of GCN. We further study the important problem, i.e., how to substantially enhance the fusion capability of GCN for classification.
\item We propose a novel adaptive multi-channel GCN framework, AM-GCN, which performs graph convolution operation over both topology and feature spaces. Combined with attention mechanism, different information can be adequately fused.
\item  Our extensive experiments on a series of benchmark data sets clearly show that AM-GCN outperforms the state-of-the-art GCNs and extracts the most correlation information from both node features and topological structures nicely for challenging classifcation tasks.
\vspace{-\topsep}\end{itemize}

The rest of the paper is organized as follows. In Section~\ref{sec:capability} we experimentally investigate the capability of GCNs in fusing node features and topology.  In Section~\ref{sec:am-gcn}, we develop AM-GCN. We report experimental results in Section~\ref{sec:exp}, and review related work in Section~\ref{sec:related-work}.  We conclude the paper in Section~\ref{sec:con}.

\vspace{-\topsep}\section{Fusion Capability of GCNs: An Experimental Investigation}\label{sec:capability}

In this section, we use two simple yet intuitive cases to examine whether the state-of-the-art GCNs can adaptively learn from node features and topological structures in graphs and fuse them sufficiently for classification tasks. The main idea is that we will clearly establish the high correlation between node label with network topology and node features, respectively, then we will check the performance of GCN on these two simple cases. A good fusion capability of GCN should adaptively extract the correlated information with the supervision of node label, providing a good result. However, if the performance drops sharply in comparison with baselines, this will demonstrate that GCN cannot adaptively extract information from node features and topological structures, even there is a high correlation between node features or topological structures with the node label.

% that the node label in the current mechanism of GCN cannot supervise to adaptively learn node embedding from network topology and node feature for classification.

\vspace{-5pt}\subsection{Case 1: Random Topology and Correlated Node Features}

We generate a random network consisting of 900 nodes, where the probability of building an edge betweean any two nodes is $0.03$. Each node has a feature vector of 50 dimensions. To generate node features, we randomly assign 3 labels to the 900 nodes, and for the nodes with the same label, we use one Gaussian distribution to generate the node features. The Gaussian distributions for the three classes of nodes  have the same covariance matrix, but three different centers far away from each other. In this data set, the node labels are highly correlated with the node features, but not the topological structures.

We apply GCN~\cite{kipf2017semi} to train this network. For each class we randomly select 20 nodes for training and another 200 nodes for testing. We carefully tune the hyper-parameters to report the best performance and avoid over smoothing. Also, we apply MLP~\cite{pal1992multilayer} to the node features only. The classification accuracies of GCN and MLP are $75.2\%$ and $100\%$, respectively.

The results meet the expectation. Since the node features are highly correlated with the node labels, MLP shows excellent performance. GCN extracts information from both the node features and the topological structures, but cannot adaptively fuse them to avoid the interference from topological structures. It cannot match the high performance of MLP.

\vspace{-5pt}\subsection{Case 2: Correlated Topology and Random Node Features}

We generate another network with 900 nodes.  This time, the node features, each of 50 dimensions, are randomly generated. For the topological structure, we employ the Stochastic Blockmodel (SBM)~\cite{karrer2011stochastic} to split nodes into 3 communities (nodes 0-299, 300-599, 600-899, respectively). Within each community, the probability of building an edge is set to $0.03$, and the probability of building an edge between nodes in different communities is set to $0.0015$. In this data set, the node labels are determined by the communities, i.e., nodes in the same community have the same label.

Again we apply GCN to this network. We also apply DeepWalk~\cite{perozzi2014deepwalk} to the topology of the network, that is, the features are ignored by DeepWalk. The classification accuracies of GCN and DeepWalk are $87\%$ and $100\%$, respectively.

DeepWalk performs well because it models network topological structures thoroughly. GCN extracts information from both the node features and the topological structures, but cannot adaptively fuse them to avoid the interference from node features. It cannot match the high performance of DeepWalk.

\textbf{Summary}. These cases show that the current fusion mechanism of GCN~\cite{kipf2017semi} is distant from optimal or even satisfactory. Even the correlation between node label with network topology or node features is very high, the current GCN cannot make full use of the supervision by node label to adaptively extract the most correlated information. However, the situation is more complex in reality, because it is hard to know whether the topology or the node features are more correlated with the final task, which prompts us to rethink the current mechanism of GCN.

\begin{figure}[t]
	\centering
	\includegraphics[width=0.48\textwidth]{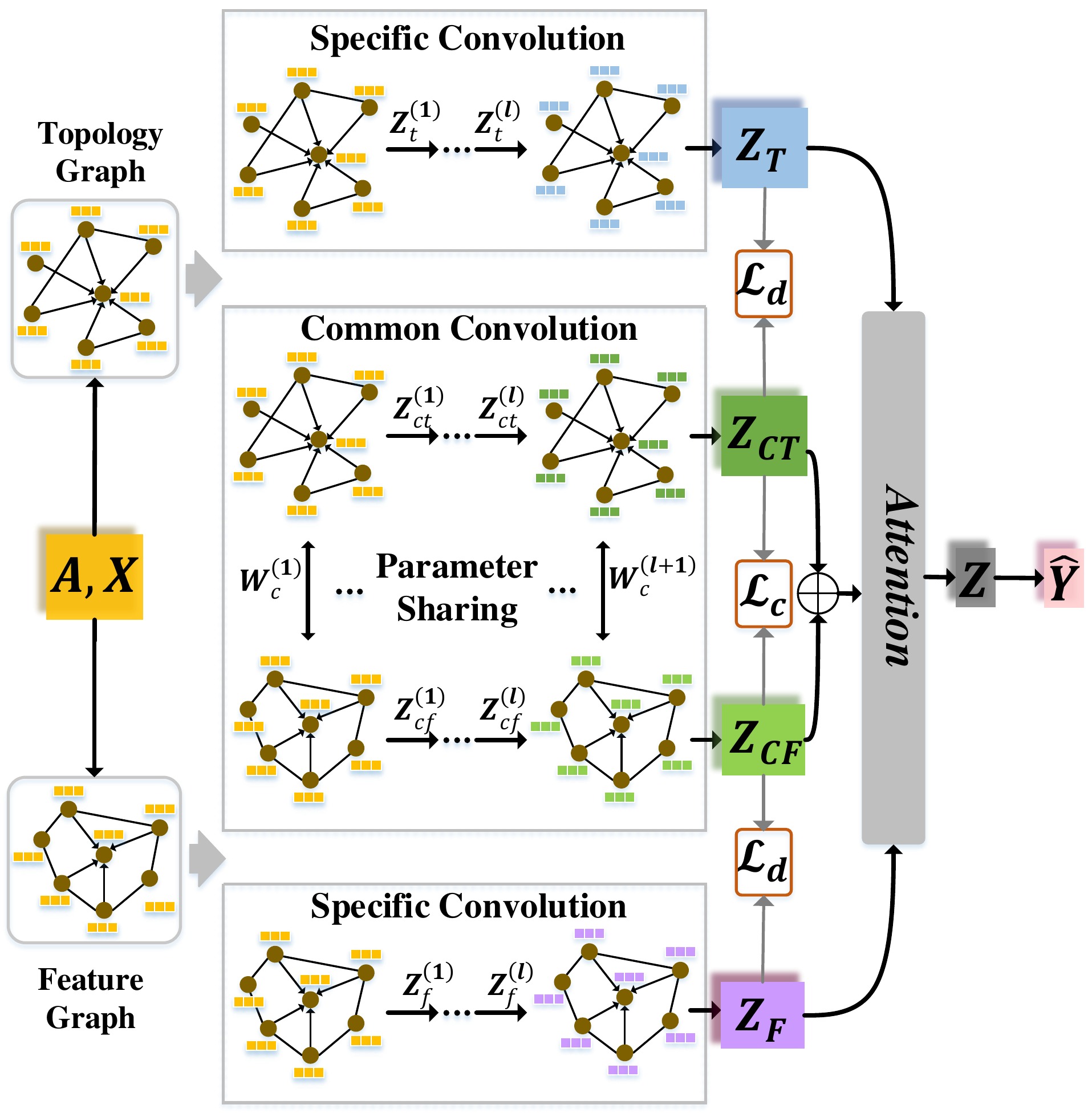}
	\setlength{\abovecaptionskip}{2pt}
	\setlength{\belowcaptionskip}{-13pt}
	\caption{The framework of AM-GCN model. Node feature \textbf{X} is to construct a feature graph. AM-GCN consists of two \textit{specific convolution modules}, one \textit{common convolution module} and the \textit{attention mechanism}.}
	\Description{AM-GCN Model.}
	\label{model}
\end{figure}

\vspace{-\topsep}\section{AM-GCN: the Proposed Model}\label{sec:am-gcn}

$\textbf{Problem Settings:}$ We focus on semi-supervised node classification in an attributed graph $G=(\textbf{A},\textbf{X})$, where $\textbf{A}\in\mathbb{R}^{n\times n}$ is the symmetric adjacency matrix with $n$ nodes and $\textbf{X}\in\mathbb{R}^{n\times d}$ is the node feature matrix, and \textit{d} is the dimension of node features. Specifically, $A_{ij}=1$ represents there is an edge between nodes $i$ and $j$, otherwise, $A_{ij}=0$. We suppose each node belongs to one out of $C$ classes.

The overall framework of AM-GCN is shown in Figure \ref{model}. The key idea is that AM-GCN permits node features to propagate not only in topology space, but also in feature space, and the most correlated information with node label should be extracted from both of these two spaces. To this end, we construct a feature graph based on node features $\textbf{X}$. Then with two specific convolution modules, $\textbf{X}$ is able to propagate over both of feature graph and topology graph to learn two specific embeddings $\textbf{Z}_F$ and $\textbf{Z}_T$, respectively. Further, considering that the information in these two spaces have common characteristics, we design a common convolution module with parameter sharing strategy to learn the common embedding $\textbf{Z}_{CF}$ and $\textbf{Z}_{CT}$, also, a consistency constraint $\mathcal{L}_{c}$ is employed to enhance the "common" property of $\textbf{Z}_{CF}$ and $\textbf{Z}_{CT}$. Besides, a disparity constraint $\mathcal{L}_{d}$ is to ensure the independence between $\textbf{Z}_F$ and $\textbf{Z}_{CF}$, as well as $\textbf{Z}_T$ and $\textbf{Z}_{CT}$. Considering that node label may be correlated with topology or feature or both, AM-GCN utilizes an attention mechanism to adaptively fuse these embeddings with the learned weights, so as to extract the most correlated information $\textbf{Z}$ for the final classification task.

\subsection{Specific Convolution Module}

Firstly, in order to capture the underlying structure of nodes in feature space, we construct a \textit{k}-nearest neighbor (\textit{k}NN) graph $G_f = (\textbf{A}_f, \textbf{X})$ based on node feature matrix \textbf{X}, where $\textbf{A}_f$ is the adjacency matrix of \textit{k}NN graph. Specifically, we first calculate the similarity matrix $\textbf{S} \in \mathbb{R}^{n\times n}$ among $n$ nodes. Actually, there are many ways to obtain \textbf{S}, and we list two popular ones here,  in which $\textbf{x}_i$ and $\textbf{x}_j$ are feature vectors of nodes $i$ and $j$:

1) \textbf{Cosine Similarity}: It uses the cosine value of the angle between two vectors to measure the similarity:
\begin{equation}
\textbf{S}_{ij} = \frac{{\textbf{x}_i}\cdot{\textbf{x}_j}}{|{\textbf{x}_i}||{\textbf{x}_j}|}.
\end{equation}

2) \textbf{Heat Kernel}: The similarity is calculated by the Eq. \eqref{kernel} where \textit{t} is the time parameter in heat conduction equation and we set $t=2$.
\begin{equation}
\textbf{S}_{ij} = e^{-\frac{\|{\textbf{x}_i}-{\textbf{x}_j}\|^2}{t}}.
\label{kernel}
\end{equation}
Here we uniformly choose the Cosine Similarity to obtain the similarity matrix $\textbf{S}$, and then we choose top $k$ similar node pairs for each node to set edges and finally get the adjacency matrix $\textbf{A}_f$.

Then with the input graph $(\textbf{A}_f, \textbf{X})$ in feature space, the \textit{l}-th layer output $\textbf{Z}^{(\textit{l})}_f$ can be represented as:
\begin{equation}
\textbf{Z}^{(\textit{l})}_f = ReLU(
\tilde{\textbf{D}}^{-\frac{1}{2}}_f
\tilde{\textbf{A}}_f
\tilde{\textbf{D}}^{-\frac{1}{2}}_f
\textbf{Z}^{(\textit{l-1})}_f
\textbf{W}^{(\textit{l})}_f),
\end{equation}
where $ \textbf{W}^{(\textit{l})}_f$ is the weight matrix of the  \textit{l}-th layer in GCN, $ReLU$ is the Relu activation function and the initial $\textbf{Z}^{(0)}_f =\textbf{X}$. Specifically, we have $\tilde{\textbf{A}}_f = \textbf{A}_f + \textbf{I}_f $ and $\tilde{\textbf{D}}_f$ is the diagonal degree matrix of $\tilde{\textbf{A}}_f$. We denote the last layer output
embedding as $\textbf{Z}_F$.
In this way, we can learn the node embedding which captures the specific information $\textbf{Z}_F$ in feature space.

As for the topology space, we have the original input graph $G_t = (\textbf{A}_t, \textbf{X}_t)$ where $ \textbf{A}_t = \textbf{A} $ and $\textbf{X}_t = \textbf{X}$. Then the learned output embedding $\textbf{Z}_T$ based on topology graph can be calculated in the same way as in feature space. Therefore, the specific information encoded in topology space can be extracted.

\subsection{\textbf{Common Convolution Module}}

In reality, the feature and topology spaces are not completely irrelevant. Basically, the node classification task, may be correlated with the information either in feature space or in topology space or in both of them, which is difficult to know beforehand. Therefore, we not only need to extract the node specific embedding in these two spaces, but also to extract the common information shared by the two spaces. In this way, it will become more flexible for the task to determine which part of information is the most correlated. To address this, we design a \textit{Common}-GCN with parameter sharing strategy to get the embedding shared in two spaces.

First, we utilize \textit{Common}-GCN to extract the node embedding $\textbf{Z}^{(\textit{l})}_{ct}$ from topology graph ($\textbf{A}_t$, $\textbf{X}$)
as follows
\begin{equation}
\label{tGCN}
  \textbf{Z}^{(\textit{l})}_{ct} = ReLU(
  \tilde{\textbf{D}}^{-\frac{1}{2}}_t
  \tilde{\textbf{A}}_t
  \tilde{\textbf{D}}^{-\frac{1}{2}}_t
  \textbf{Z}^{(\textit{l-1})}_{ct}
  \textbf{W}^{(\textit{l})}_c),
\end{equation}
where $ \textbf{W}^{(\textit{l})}_c$ is the \textit{l}-th layer weight matrix of \textit{Common}-GCN and $\textbf{Z}^{(\textit{l-1})}_{ct}$ is the node embedding in the $(l-1)$th layer and $\textbf{Z}^{(0)}_{ct}=\textbf{X}$.
When utilizing \textit{Common}-GCN to learn the node embedding from feature graph ($\textbf{A}_f$, $\textbf{X}$), in order to extract the shared information, we share the same weight matrix $\textbf{W}^{(\textit{l})}_c$ for every layer of \textit{Common}-GCN as follows:
\begin{equation}
\label{fGCN}
  \textbf{Z}^{(\textit{l})}_{cf} = ReLU(
  \tilde{\textbf{D}}^{-\frac{1}{2}}_f
  \tilde{\textbf{A}}_f
  \tilde{\textbf{D}}^{-\frac{1}{2}}_f
  \textbf{Z}^{(\textit{l-1})}_{cf}
  \textbf{W}^{(\textit{l})}_c),
\end{equation}
where $\textbf{Z}^{(\textit{l})}_{cf}$ is the l-layer output embedding and $\textbf{Z}^{(0)}_{cf}=\textbf{X}$.
The shared weight matrix can filter out the shared characteristics from two spaces. According to different input graphs, we can get two output embedding $\textbf{Z}_{CT}$ and $\textbf{Z}_{CF}$ and the common embedding $\textbf{Z}_C$ of the two spaces is:
\begin{equation}
\textbf{Z}_C = (\textbf{Z}_{CT} + \textbf{Z}_{CF})/2.
\end{equation}

\subsection{Attention Mechanism}

Now we have two specific embeddings $\textbf{Z}_T$ and $\textbf{Z}_F$, and one common embedding $\textbf{Z}_C$. Considering the node label can be correlated with one of them or even their combinations, we use the attention mechanism $\textit{att}(\textbf{Z}_{T}, \textbf{Z}_{C}, \textbf{Z}_{F})$ to learn their corresponding importance $(\bm{\alpha}_t, \bm{\alpha}_c, \bm{\alpha}_f)$ as follows:
\begin{equation}
(\bm{\alpha}_t, \bm{\alpha}_c, \bm{\alpha}_f) = \textit{att} (\textbf{Z}_{T}, \textbf{Z}_{C}, \textbf{Z}_{F}),
\end{equation}
here $\bm{\alpha}_t, \bm{\alpha}_c, \bm{\alpha}_f \in \mathbb{R}^{n\times1}$ indicate the attention values
of $n$ nodes with embeddings $\textbf{Z}_{T}, \textbf{Z}_{C}, \textbf{Z}_{F}$, respectively.

Here we focus on node $i$, where its embedding in $\textbf{Z}_{T}$ is $\textbf{z}_{T}^i \in \mathbb{R}^{1\times h} $ (i.e., the $i$-th row of $\textbf{Z}_T$).
We firstly transform the embedding through a nonlinear transformation, and then use one shared attention vector $\textbf{q} \in \mathbb{R}^{h'\times 1}$ to get the attention value $\omega_{T}^i$ as follows:
\begin{equation}
\omega_T^i = \textbf{q}^T \cdot tanh(\textbf{W} \cdot (\textbf{z}_T^i)^T + \textbf{b}).
\end{equation}
Here $\textbf{W} \in \mathbb{R}^{h'\times h}$ is the weight matrix and $\textbf{b}\in \mathbb{R}^{h'\times 1}$ is the bias vector. Similarly, we can get the attention values $\omega_{C}^i$
and $\omega_{F}^i$ for node $i$ in embedding matrices $\textbf{Z}_C$ and $\textbf{Z}_F$, respectively.
We then normalize the attention values $\omega_{T}^i, \omega_{C}^i, \omega_{F}^i$ with softmax function to get
 the final weight:
\begin{equation}
\alpha_T^i =softmax(\omega_T^i) =  \frac{exp(\omega_T^i)}{exp(\omega_T^i)+exp(\omega_C^i)+exp(\omega_F^i)}.
\end{equation}
Larger $\alpha_T^i$ implies the corresponding embedding is more important. Similarly, $\alpha_C^i =softmax(\omega_C^i)$
and $\alpha_F^i =softmax(\omega_F^i)$.
For all the $n$ nodes, we have the learned weights $\bm{\alpha}_t=[\alpha_T^i], \bm{\alpha}_c=[\alpha_C^i], \bm{\alpha}_f=[\alpha_F^i] \in \mathbb{R}^{n\times1}$, and denote $\bm{\alpha_T} = diag(\bm{\alpha}_t)$, $\bm{\alpha_C} = diag(\bm{\alpha}_c)$ and  $\bm{\alpha_F} = diag(\bm{\alpha}_f)$. Then we combine these three embeddings to obtain the final embedding \textbf{Z} :
\begin{equation}
\label{output}
\textbf{Z} = \bm{\alpha_T} \cdot \textbf{Z}_T + \bm{\alpha_C} \cdot \textbf{Z}_C + \bm{\alpha_F} \cdot \textbf{Z}_F.
\end{equation}

\subsection{Objective Function}

\subsubsection{\textbf{Consistency Constraint}}

For the two output embeddings $\textbf{Z}_{CT}$ and $\textbf{Z}_{CF}$ of \textit{Common}-GCN, despite the \textit{Common}-GCN has the shared weight matrix, here we design a consistency constraint to further enhance their commonality.

Firstly, we use $L_2$-normalization to normalize the embedding matrix as $\textbf{Z}_{CTnor}$, $\textbf{Z}_{CFnor}$.
Then, the two normalized matrix can be used to capture the similarity of $n$ nodes as $\textbf{S}_{T}$ and $\textbf{S}_{F}$ as follows:
\begin{equation}
\begin{aligned}
\textbf{S}_{T} &= \textbf{Z}_{CTnor}^{} \cdot \textbf{Z}_{CTnor}^T,\\
\textbf{S}_{F} &= \textbf{Z}_{CFnor}^{} \cdot \textbf{Z}_{CFnor}^T.
\end{aligned}
\end{equation}

The consistency implies that the two similarity matrices should be similar, which gives rise to the following constraint:
\begin{equation}
\mathcal{L}_{c} = \|\textbf{S}_{T}-\textbf{S}_{F}\|_F^2.
\end{equation}

\subsubsection{\textbf{Disparity Constraint}}
Here because embeddings $\textbf{Z}_T$ and $\textbf{Z}_{CT}$ are learned from the same graph $G_t = (\textbf{A}_t, \textbf{X}_t)$,
to ensure they can capture different information, we employ the Hilbert-Schmidt Independence Criterion (HSIC)~\cite{song2007supervised}, a simple but effective measure of independence, to enhance the disparity of these two embeddings. Due to its simplicity and neat theoretical properties, HSIC has been applied to several machine learning tasks~\cite{niu2010multiple, gretton2005measuring}. Formally, the HSIC constraint of $\textbf{Z}_T$ and $\textbf{Z}_{CT}$ is defined as:
\begin{equation}
HSIC(\textbf{Z}_{T}, \textbf{Z}_{CT}) = (n-1)^{-2}tr(\textbf{R}\textbf{K}_{T}\textbf{R}\textbf{K}_{CT}),
\end{equation}
where $\textbf{K}_{T}$ and $\textbf{K}_{CT}$ are the Gram matrices with $k_{T,ij}=k_T(\textbf{z}_{T}^i,\textbf{z}_{T}^j)$ and $k_{CT,ij}=k_{CT}(\textbf{z}_{CT}^i,\textbf{z}_{CT}^j)$. And $\textbf{R} = \textbf{I} - \frac{1}{n}ee^T$, where \textbf{I} is an identity matrix and $e$ is an all-one column vector. In our implementation, we use the inner product kernel function for $\textbf{K}_{T}$ and $\textbf{K}_{CT}$.

Similarly, considering the embeddings $\textbf{Z}_F$ and $\textbf{Z}_{CF}$ are also learned from the same graph $(\textbf{A}_f, \textbf{X})$, their disparity should also be enhanced by HSIC:
\begin{equation}
HSIC(\textbf{Z}_{F}, \textbf{Z}_{CF}) = (n-1)^{-2}tr(\textbf{R}\textbf{K}_{F}\textbf{R}\textbf{K}_{CF}).
\end{equation}

Then we set the disparity constraint as $\mathcal{L}_{d}$ where:
\begin{equation}
\mathcal{L}_{d} = HSIC(\textbf{Z}_{T}, \textbf{Z}_{CT}) + HSIC(\textbf{Z}_{F}, \textbf{Z}_{CF}).
\end{equation}

\begin{table}[htbp]
	\caption{The statistics of the datasets}
	\label{dataset}
	\setlength{\tabcolsep}{0.5mm}{
		\begin{tabular}{lcccccc}
			\hline
			Dataset&Nodes&Edges&Classes&Features&Training&Test\\
			\hline
			Citeseer&3327&4732&6&3703&120/240/360&1000\\
			UAI2010&3067&28311&19&4973&380/760/1140&1000\\
			ACM&3025&13128&3&1870&60/120/180&1000\\
			BlogCatalog&5196&171743&6&8189&120/240/360&1000\\
			Flickr&7575&239738&9&12047&180/360/540&1000\\
			CoraFull&19793&65311&70&8710&1400/2800/4200&1000\\
			\hline
	\end{tabular}}
\end{table}

\subsubsection{\textbf{Optimization Objective}}

We use the output embedding \textbf{Z} in Eq. \eqref{output} for semi-supervised multi-class classification with a linear transformation and a softmax function. Denote the class predictions for $n$ nodes as $\hat{\textbf{Y}}=[\hat{y}_{ic}]\in \mathbb{R}^{n\times C}$ where $\hat{y}_{ic}$ is the probability of node \textit{i} belonging to class \textit{c}. Then the $\hat{\textbf{Y}}$ can be calculated in the following way:
\begin{equation}
\hat{\textbf{Y}} = softmax(\textbf{W} \cdot \textbf{Z} + \textbf{b}),
\end{equation}
where $softmax(x)=\frac{\exp(x)}{\Sigma_{c=1}^{C}exp(x_c)}$ is actually a normalizer across all classes.

Suppose the training set is $L$, for each $l\in L$ the real label is $\textbf{Y}_l$ and the predicted label is $\hat{\textbf{Y}}_l$. Then the cross-entropy loss for node classification over all training nodes is represented as $\mathcal{L}_{t}$ where:
\begin{equation}
\mathcal{L}_{t} = -\sum\nolimits_{l\in L}\sum\nolimits_{i=1}^{C}\textbf{Y}_{li}\rm{ln}\hat{\textbf{Y}}_\textit{li}.
\end{equation}

Combining the node classification task and constraints, we have the following overall objective function:
\begin{equation}
\label{final_target}
\mathcal{L} = \mathcal{L}_{t}+\gamma\mathcal{L}_{c}+\beta\mathcal{L}_{d},
\end{equation}
where $\gamma$ and $\beta$ are parameters of the consistency and disparity constraint terms. With the guide of labeled data, we can optimize the proposed model via back propagation and learn the embedding of nodes for classification.

\section{Experiments}\label{sec:exp}
\subsection{Experimental Setup}

\noindent{\bfseries Datasets}
\quad Our proposed AM-GCN is evaluated on six real world datasets which are summarized in Table \ref{dataset}, moreover, we provide all the data websites in the supplement for reproducibility.

\begin{itemize}
\item \textbf{Citeseer}~\cite{kipf2017semi}: Citeseer is a research paper citation network, where nodes are publications and edges are citation links. Node attributes are bag-of-words representations of the papers and all nodes are divided into six areas.
\item \textbf{UAI2010}~\cite{wang2018a}: We use this dataset with 3067 nodes and 28311 edges which has been tested in graph convolutional networks for community detection in~\cite{wang2018a}.
\item \textbf{ACM}~\cite{wang2019heterogeneous}: This network is extracted from ACM dataset where nodes represent papers and there is an edge between two papers if they have the same author. All the papers are divided into 3 classes (\textit{Database}, \textit{Wireless Communication}, \textit{DataMining}). The features are the bag-of-words representations of paper keywords.
\item \textbf{BlogCatalog}~\cite{meng2019co}: It is a social network with bloggers and their social relationships from the BlogCatalog website. Node attributes are constructed by the keywords of user profiles, and the labels represent the topic categories provided by the authors, and all nodes are divided into 6 classes.
\item \textbf{Flickr}~\cite{meng2019co}:  Flickr is an image and video hosting website, where users interact with each other via photo sharing. It is a social network where nodes represent users and edges represent their relationships, and all the nodes are divided into 9 classes according to interest groups of users.

\item \textbf{CoraFull}~\cite{bojchevski2018deep}:  This is the larger version of the well-known citation network Cora dataset, where nodes represent papers and edges represents their citations, and the nodes are labeled based on the paper topics.

\end{itemize}

\noindent{\bfseries Baselines} \quad We compare AM-GCN with two types of state-of-the-art methods, covering two network embedding algorithms and six graph neural network based methods. Moreover, we provide all the code websites in the supplement for reproducibility.

\begin{itemize}
\item \textbf{DeepWalk}~\cite{perozzi2014deepwalk} is a network embedding method which uses random walk to obtain contextual information and uses skip-gram algorithm to learn network representations.
\item \textbf{LINE}~\cite{tang2015line} is a large-scale network embedding method preserving first-order and second-order proximity of the network separately. Here we use LINE (1st+2nd).
\item \textbf{Chebyshev}~\cite{defferrard2016convolutional} is a GCN-based method utilizing Chebyshev filters.
\item \textbf{GCN}~\cite{kipf2017semi} is a semi-supervised graph convolutional network model which learns node representations by aggregating information from neighbors.
\item \textbf{\textit{k}NN-GCN}. For comparison, instead of traditional topology graph, we use the sparse \textit{k}-nearest neighbor graph calculated from feature matrix as the input graph of GCN and represent it as \textit{k}NN-GCN.
\item \textbf{GAT}~\cite{ve2018graph} is a graph neural network model using attention mechanism to aggregate node features.
\item \textbf{DEMO-Net}~\cite{wu2019demo} is a degree-specific graph neural network for node classification.
\item \textbf{MixHop}~\cite{abu-el-haija2019mixhop} is a GCN-based method which mixes the feature representations of higher-order neighbors in one graph convolution layer.
\end{itemize}

\begin{table*}
	\caption{Node classification results(\%). (Bold: best; Underline: runner-up.) }%$-$ means out of memory.}
	\label{node classification}
	\begin{tabular}{c|c|c||cccccccc|c}
		\hline
		Datasets&Metrics&L/C &DeepWalk&LINE&Chebyshev&GCN&\textit{k}NN-GCN&GAT&DEMO-Net&MixHop&AM-GCN\\
		\hline
		\multirow{6}{*}{Citeseer}&
		\multirow{3}{*}{ACC}
		&20&43.47&32.71&69.80&70.30&61.35&\underline{72.50}&69.50&71.40&\textbf{73.10}\\
		& &40&45.15&33.32&71.64&\underline{73.10}&61.54&73.04&70.44&71.48&\textbf{74.70}\\
		& &60&48.86&35.39&73.26&74.48&62.38&\underline{74.76}&71.86&72.16&\textbf{75.56}\\
		\cline{2-12}
		&\multirow{3}{*}{F1}
		&20&38.09&31.75&65.92&67.50&58.86&\underline{68.14}&67.84&66.96&\textbf{68.42}\\
		& &40&43.18&32.42&68.31&\underline{69.70}&59.33&69.58&66.97&67.40&\textbf{69.81}\\
		& &60&48.01&34.37&70.31&\underline{71.24}&60.07&\textbf{71.60}&68.22&69.31&70.92\\
		\hline
		\multirow{6}{*}{UAI2010}&
		\multirow{3}{*}{ACC}
		&20&42.02&43.47&50.02&49.88&\underline{66.06}&56.92&23.45&61.56&\textbf{70.10}\\
		& &40&51.26&45.37&58.18&51.80&\underline{68.74}&63.74&30.29&65.05&\textbf{73.14}\\
		& &60&54.37&51.05&59.82&54.40&\underline{71.64}&68.44&34.11&67.66&\textbf{74.40}\\
		\cline{2-12}
		&\multirow{3}{*}{F1}
		&20&32.93&37.01&33.65&32.86&\underline{52.43}&39.61&16.82&49.19&\textbf{55.61}\\
		& &40&46.01&39.62&38.80&33.80&\underline{54.45}&45.08&26.36&53.86&\textbf{64.88}\\
		& &60&44.43&43.76&40.60&34.12&54.78&48.97&29.05&\underline{56.31}&\textbf{65.99}\\
		\hline
		\multirow{6}{*}{ACM}&
		\multirow{3}{*}{ACC}
		&20&62.69&41.28&75.24&\underline{87.80}&78.52&87.36&84.48&81.08&\textbf{90.40}\\
		& &40&63.00&45.83&81.64&\underline{89.06}&81.66&88.60&85.70&82.34&\textbf{90.76}\\
		& &60&67.03&50.41&85.43&\underline{90.54}&82.00&90.40&86.55&83.09&\textbf{91.42}\\
		\cline{2-12}
		&\multirow{3}{*}{F1}
		&20&62.11&40.12&74.86&\underline{87.82}&78.14&87.44&84.16&81.40&\textbf{90.43}\\
		& &40&61.88&45.79&81.26&\underline{89.00}&81.53&88.55&84.83&81.13&\textbf{90.66}\\
		& &60&66.99&49.92&85.26&\underline{90.49}&81.95&90.39&84.05&82.24&\textbf{91.36}\\
		\hline
		\multirow{6}{*}{BlogCatalog}&
		\multirow{3}{*}{ACC}
		&20&38.67&58.75&38.08&69.84&\underline{75.49}&64.08&54.19&65.46&\textbf{81.98}\\
		& &40&50.80&61.12&56.28&71.28&\underline{80.84}&67.40&63.47&71.66&\textbf{84.94}\\
		& &60&55.02&64.53&70.06&72.66&\underline{82.46}&69.95&76.81&77.44&\textbf{87.30}\\
		\cline{2-12}
		&\multirow{3}{*}{F1}
		&20&34.96&57.75&33.39&68.73&\underline{72.53}&63.38&52.79&64.89&\textbf{81.36}\\
		& &40&48.61&60.72&53.86&70.71&\underline{80.16}&66.39&63.09&70.84&\textbf{84.32}\\
		& &60&53.56&63.81&68.37&71.80&\underline{81.90}&69.08&76.73&76.38&\textbf{86.94}\\
		\hline
		\multirow{6}{*}{Flickr}&
		\multirow{3}{*}{ACC}
		&20&24.33&33.25&23.26&41.42&\underline{69.28}&38.52&34.89&39.56&\textbf{75.26}\\
		& &40&28.79&37.67&35.10&45.48&\underline{75.08}&38.44&46.57&55.19&\textbf{80.06}\\
		& &60&30.10&38.54&41.70&47.96&\underline{77.94}&38.96&57.30&64.96&\textbf{82.10}\\
		\cline{2-12}
		&\multirow{3}{*}{F1}
		&20&21.33&31.19&21.27&39.95&\underline{70.33}&37.00&33.53&40.13&\textbf{74.63}\\
		& &40&26.90&37.12&33.53&43.27&\underline{75.40}&36.94&45.23&56.25&\textbf{79.36}\\
		& &60&27.28&37.77&40.17&46.58&\underline{77.97}&37.35&56.49&65.73&\textbf{81.81}\\
		\hline
		\multirow{6}{*}{CoraFull}&
		\multirow{3}{*}{ACC}
		&20&29.33&17.78&53.38&56.68&41.68&\underline{58.44}&54.50&47.74&\textbf{58.90}\\
		& &40&36.23&25.01&58.22&60.60&44.80&\underline{62.98}&60.28&57.20&\textbf{63.62}\\
		& &60&40.60&29.65&59.84&62.00&46.68&\underline{64.38}&61.58&60.18&\textbf{65.36}\\
		\cline{2-12}
		&\multirow{3}{*}{F1}
		&20&28.05&18.24&47.59&52.48&37.15&\underline{54.44}&50.44&45.07&\textbf{54.74}\\
		& &40&33.29&25.43&53.47&55.57&40.42&\underline{58.30}&56.26&53.55&\textbf{59.19}\\
		& &60&37.95&30.87&54.15&56.24&43.22&\underline{59.61}&57.26&56.40&\textbf{61.32}\\
		\hline
		
	\end{tabular}%}
\end{table*}

\noindent{\bfseries Parameters Setting}
\quad To more comprehensively evaluate our model, we select three label rates for training set (i.e., 20, 40, 60 labeled nodes per class) and choose 1000 nodes as the test set. All baselines are initialized with same parameters suggested by their papers and we also further carefully turn parameters to get optimal performance. For our model, we train three 2-layer GCNs with the same hidden layer dimension (\textit{nhid1}) and the same output dimension (\textit{nhid2}) simultaneously, where \textit{nhid1} $\in\{512, 768\}$ and \textit{nhid2} $\in\{32, 128, 256\}$. We use $0.0001\thicksim0.0005$ learning rate with Adam optimizer. In addition, the dropout rate is 0.5, weight decay $\in\{5e-3, 5e-4\}$ and \textit{k} $\in\{2\ldots10\}$ for \textit{k}-nearest neighbor graph. The coefficient of consistency constraint and disparity constraints are searched in $\{0.01, 0.001, 0.0001\}$ and $\{1e-10, 5e-9, 1e-9, 5e-8 , 1e-8\}$. For all methods, we run 5 times with the same partition and report the average results. And we use Accuracy (ACC) and macro F1-score (F1) to evaluate performance of models. For the reproducibility, we provide the specific parameter values in the supplement (Section~\ref{sec:parameters}).

\begin{figure*}[t]
	\centering
	\vspace{-5pt}
	\subfigure{
		\includegraphics[width=0.27\linewidth]{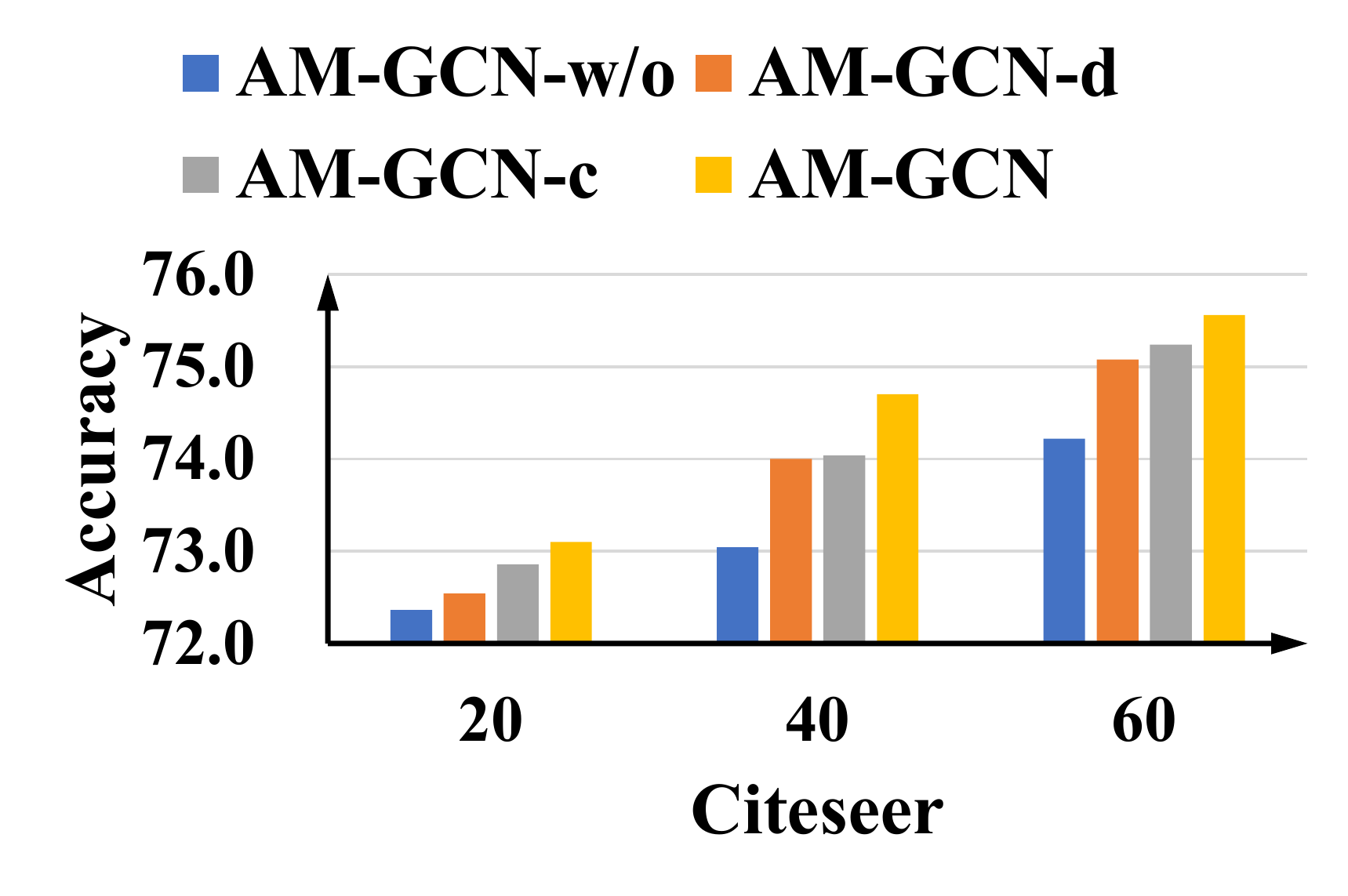}
		\label{citeseer}
	}
	\vspace{-5pt}
	\subfigure{
		\includegraphics[width=0.27\linewidth]{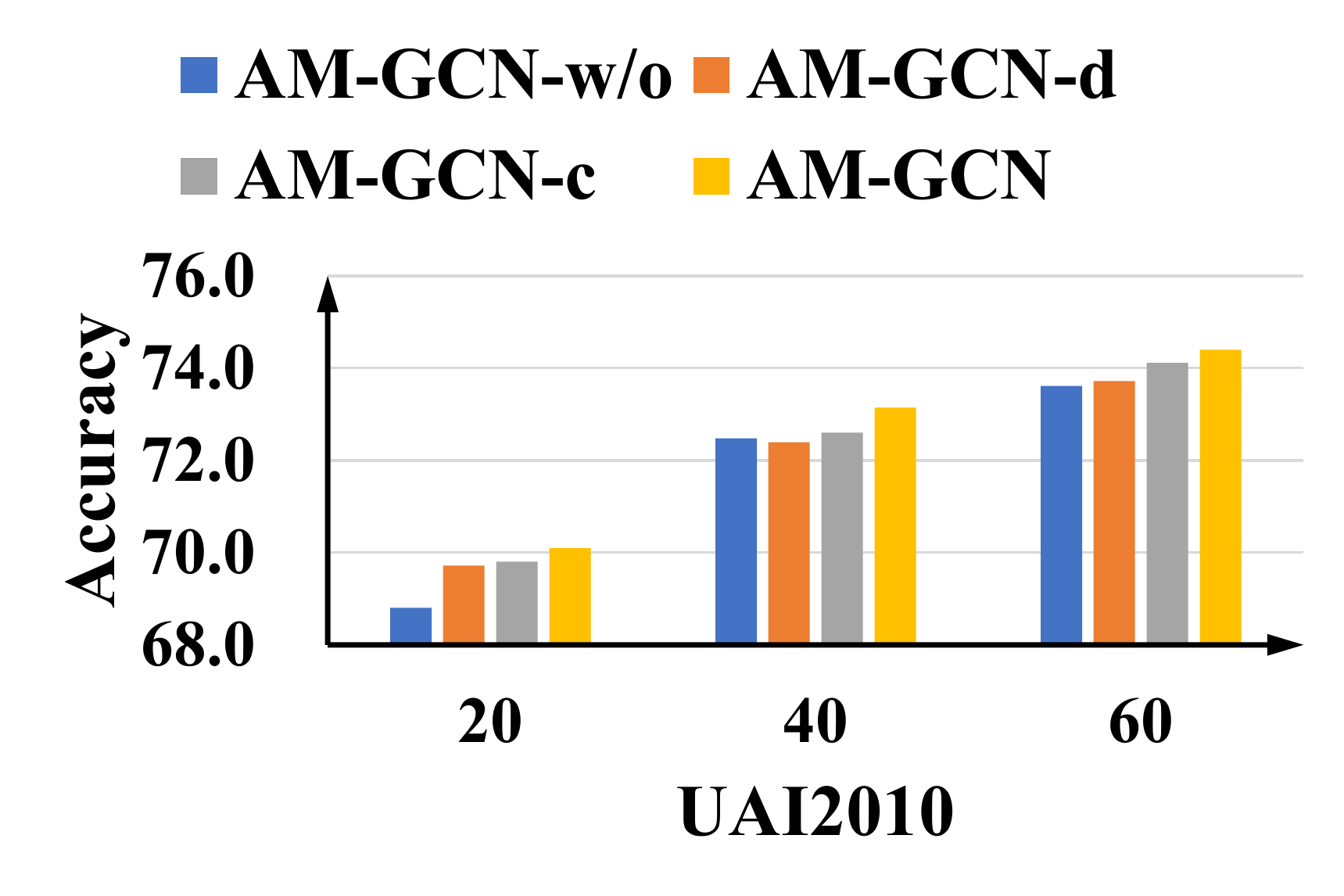}
	}
	\subfigure{
		\includegraphics[width=0.27\linewidth]{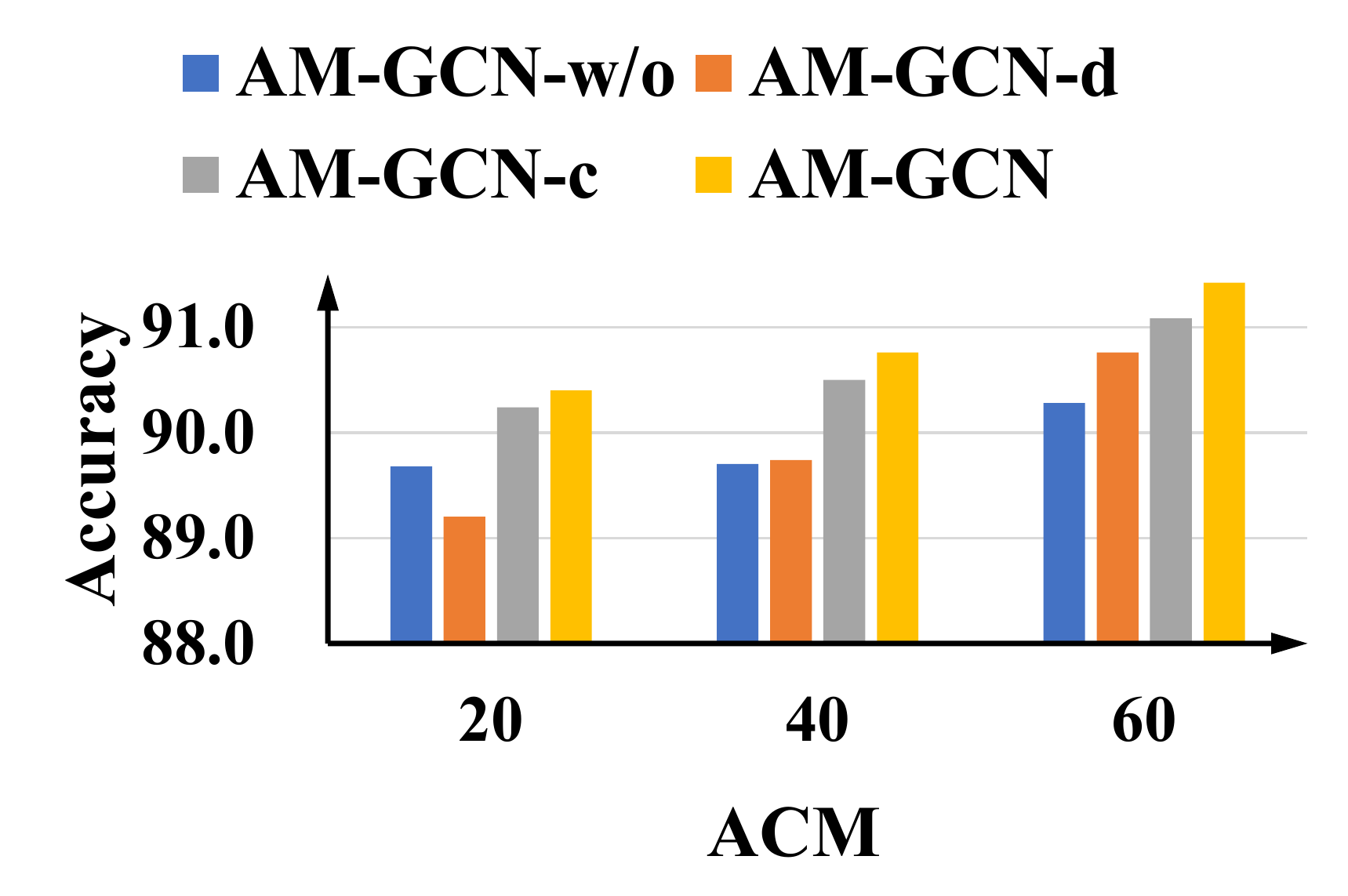}
	}
	\vspace{-5pt}
	\subfigure{
		\includegraphics[width=0.27\linewidth]{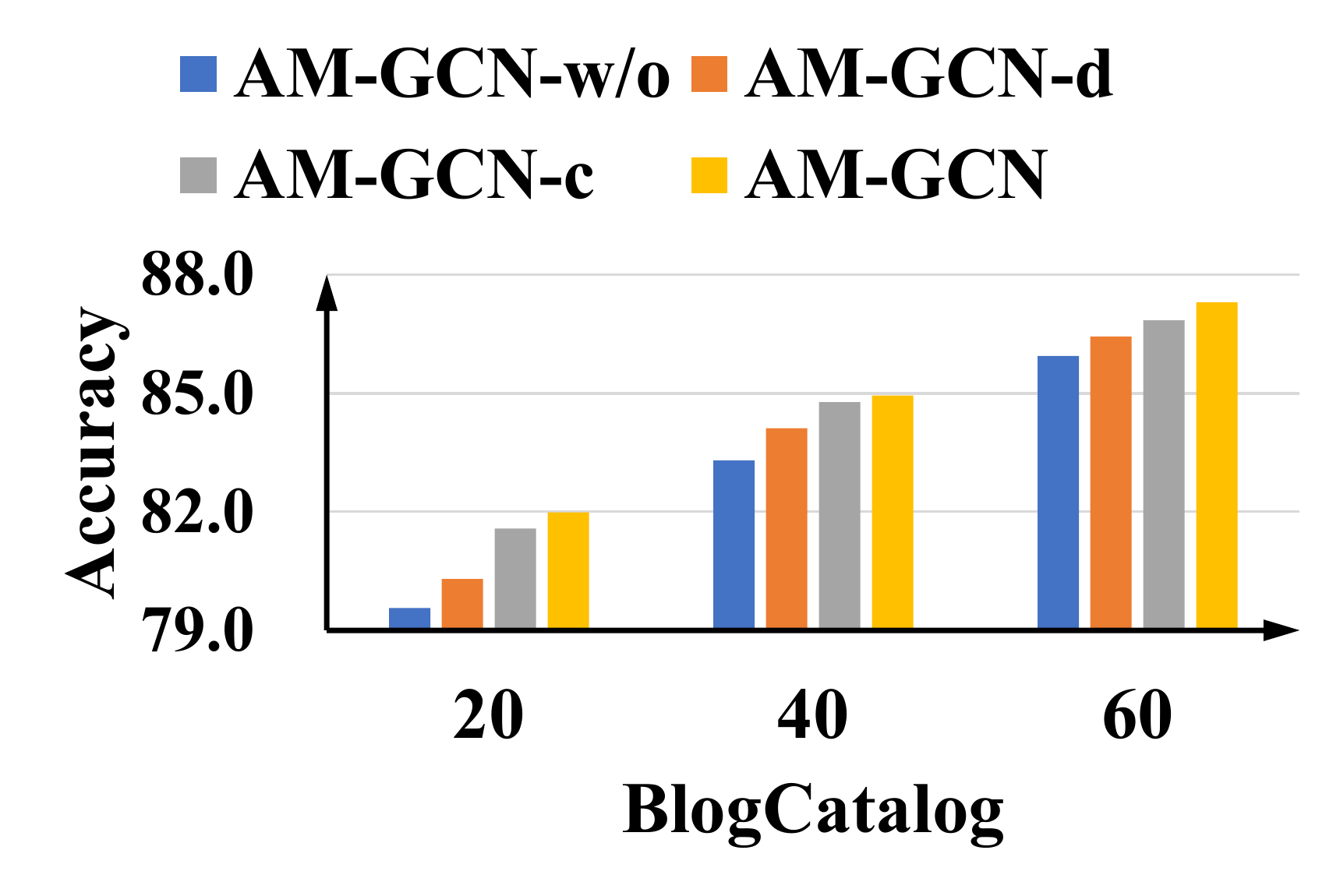}	
	}
	\subfigure{
		\includegraphics[width=0.27\linewidth]{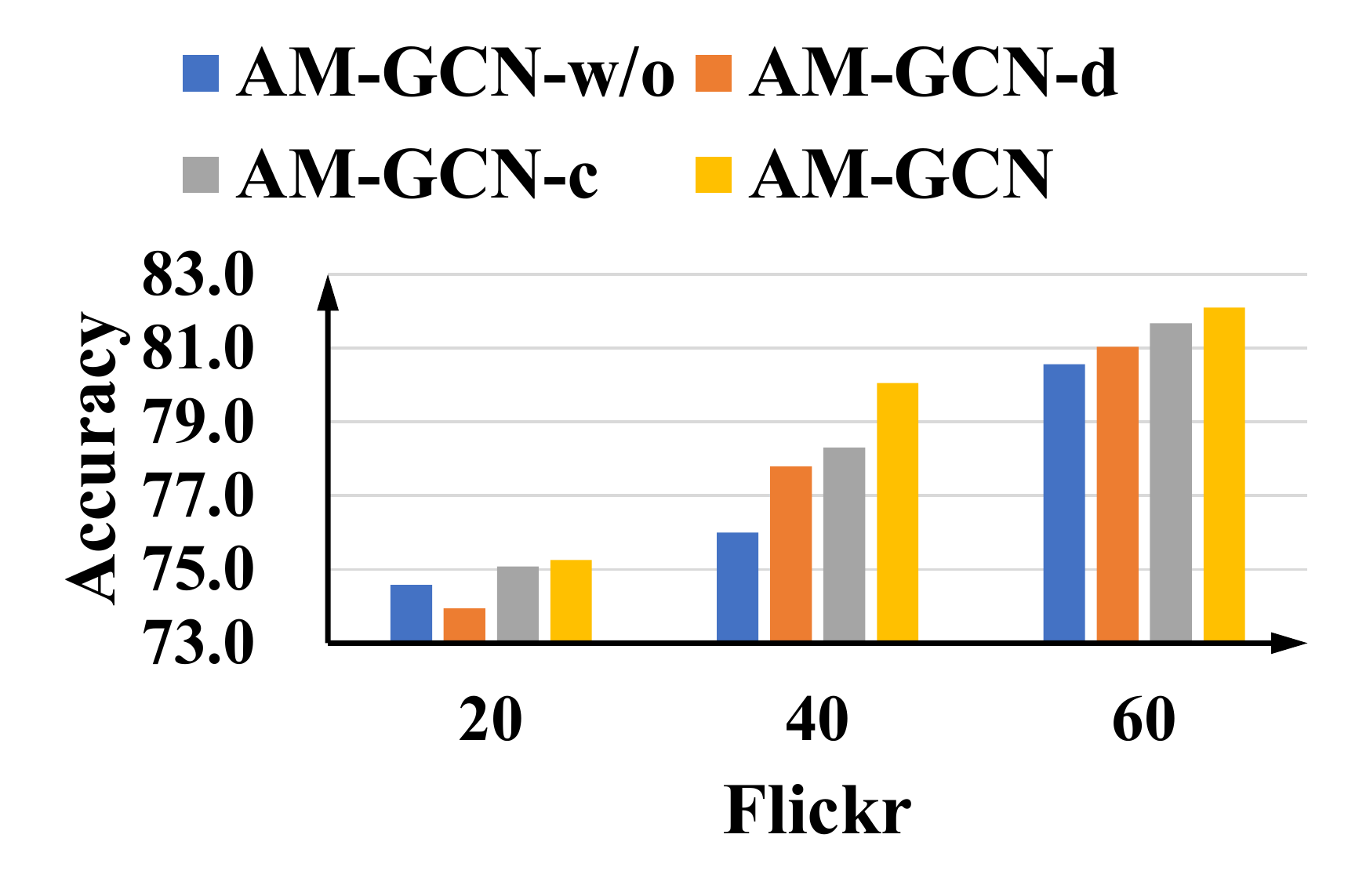}
	}
	\subfigure{
		\includegraphics[width=0.27\linewidth]{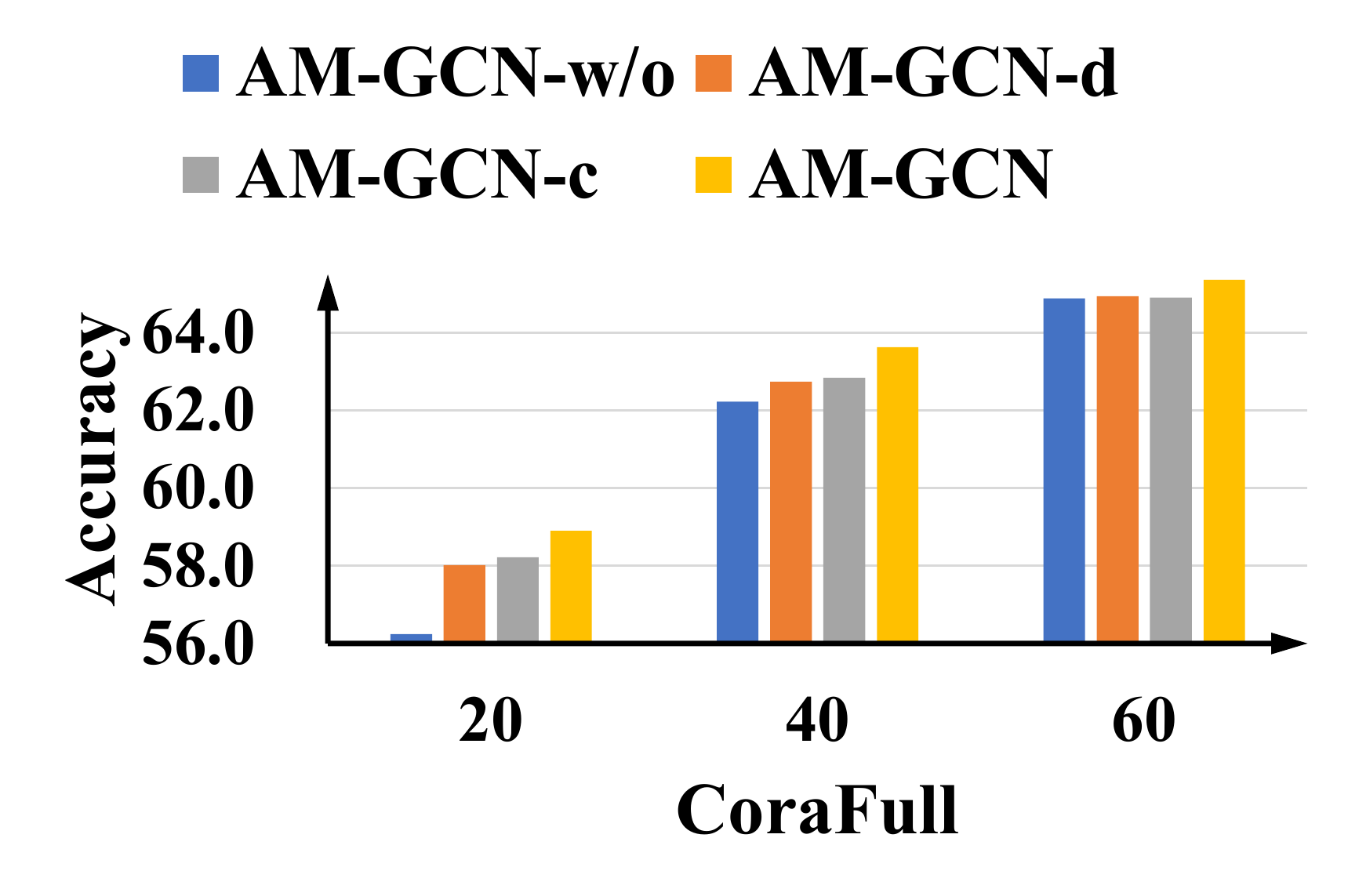}}
	\vspace{-5pt}
	\caption{The results(\%) of AM-GCN and its variants on six datasets.}
	\label{variants}
\end{figure*}

\begin{figure*}[htbp]
	\vspace{-15pt}
	\centering
	\subfigure[\textbf{DeepWalk}]{
		\includegraphics[width=0.23\textwidth]{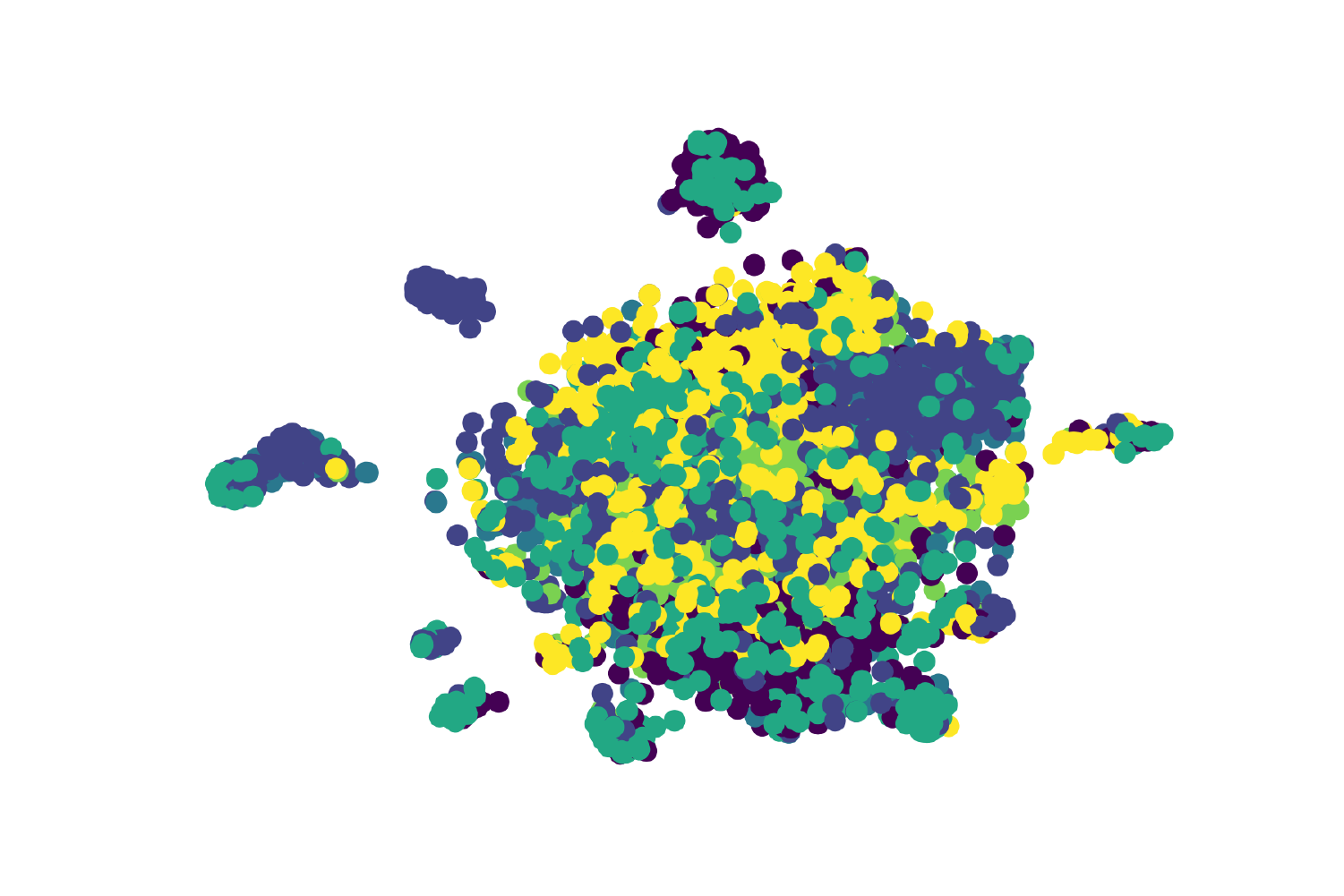}
	}
	\subfigure[\textbf{GCN}]{
		\includegraphics[width=0.23\textwidth]{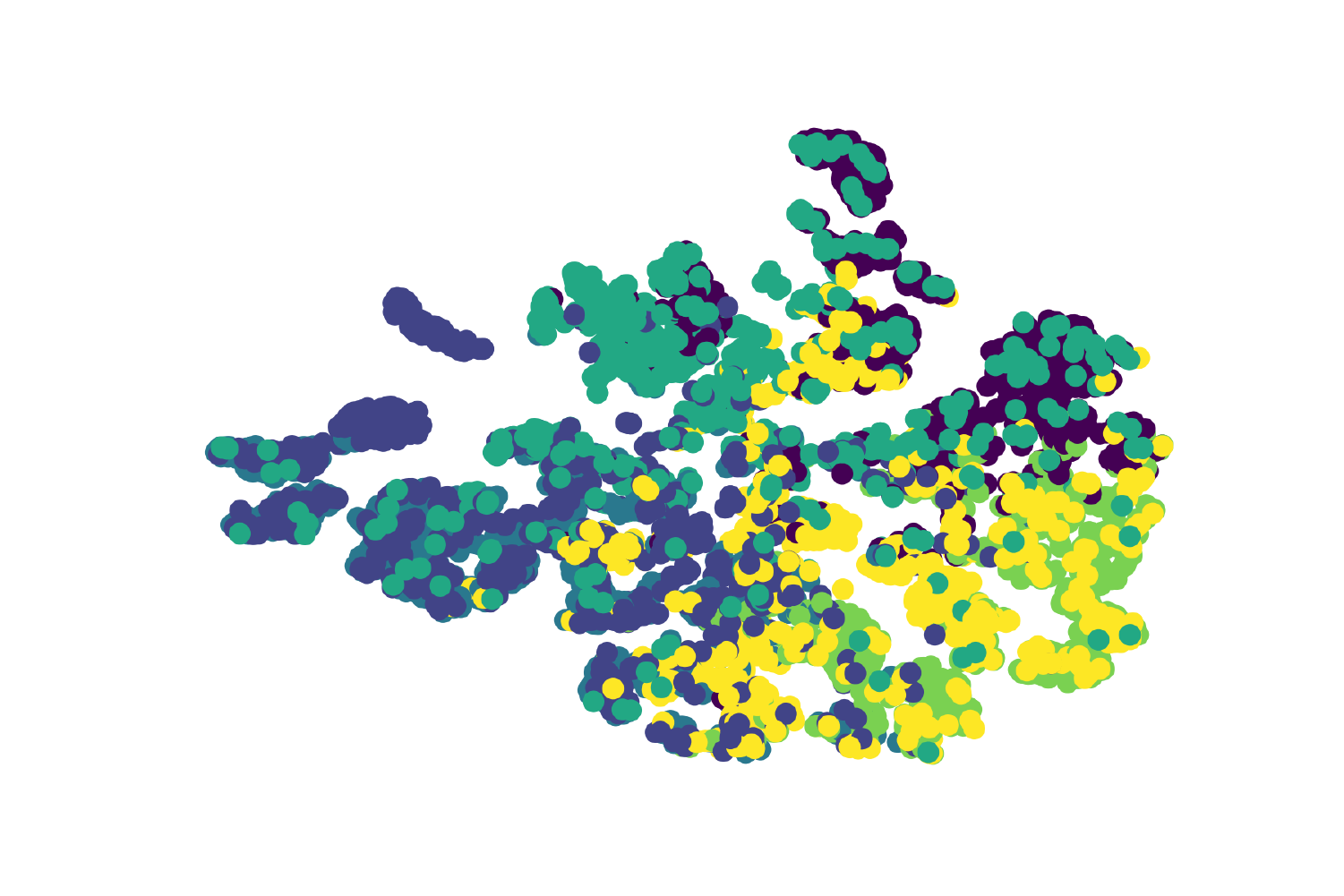}
	}
	\subfigure[\textbf{GAT}]{
		\includegraphics[width=0.23\textwidth]{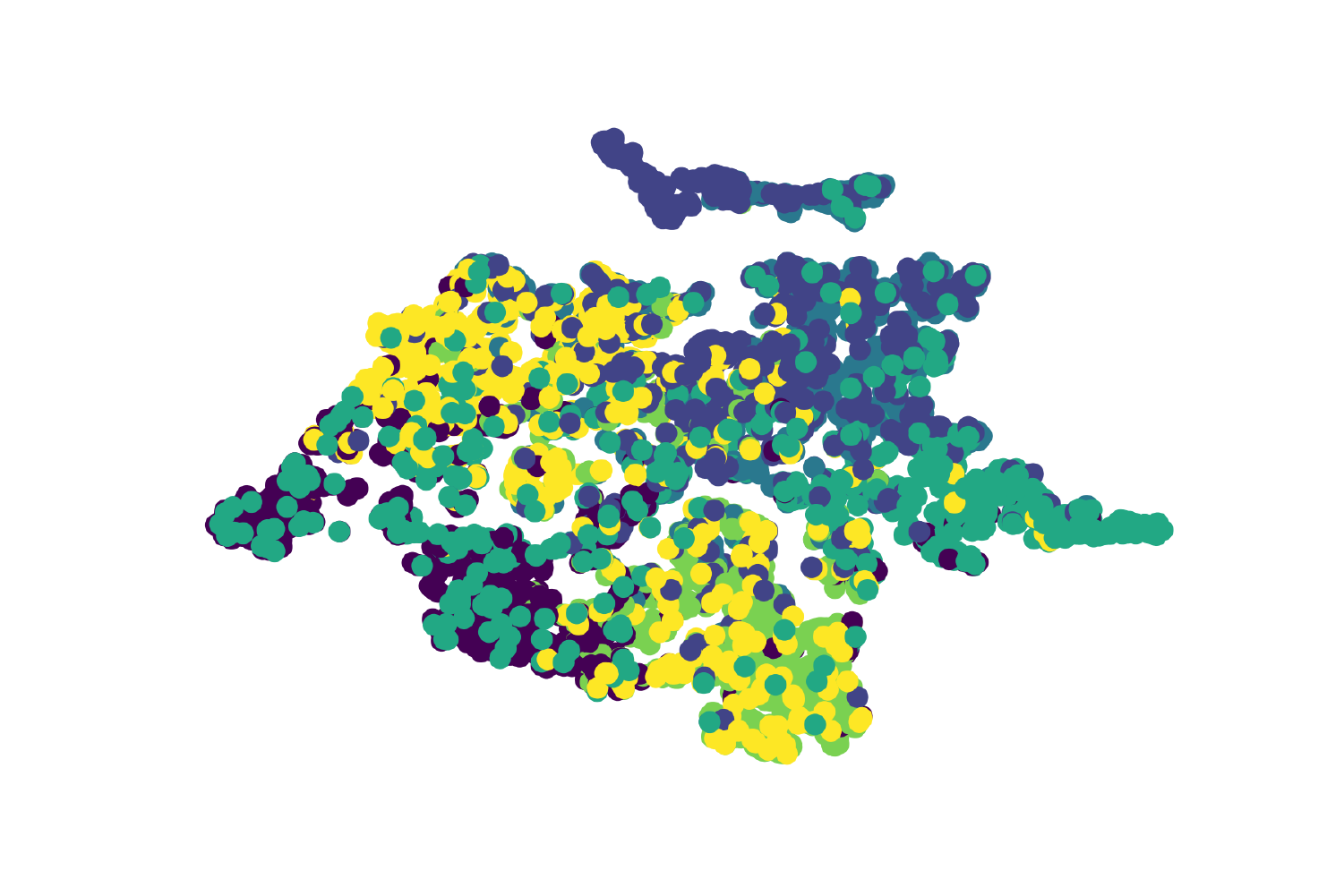}
	}
	\subfigure[\textbf{AM-GCN}]{
		\includegraphics[width=0.23\textwidth]{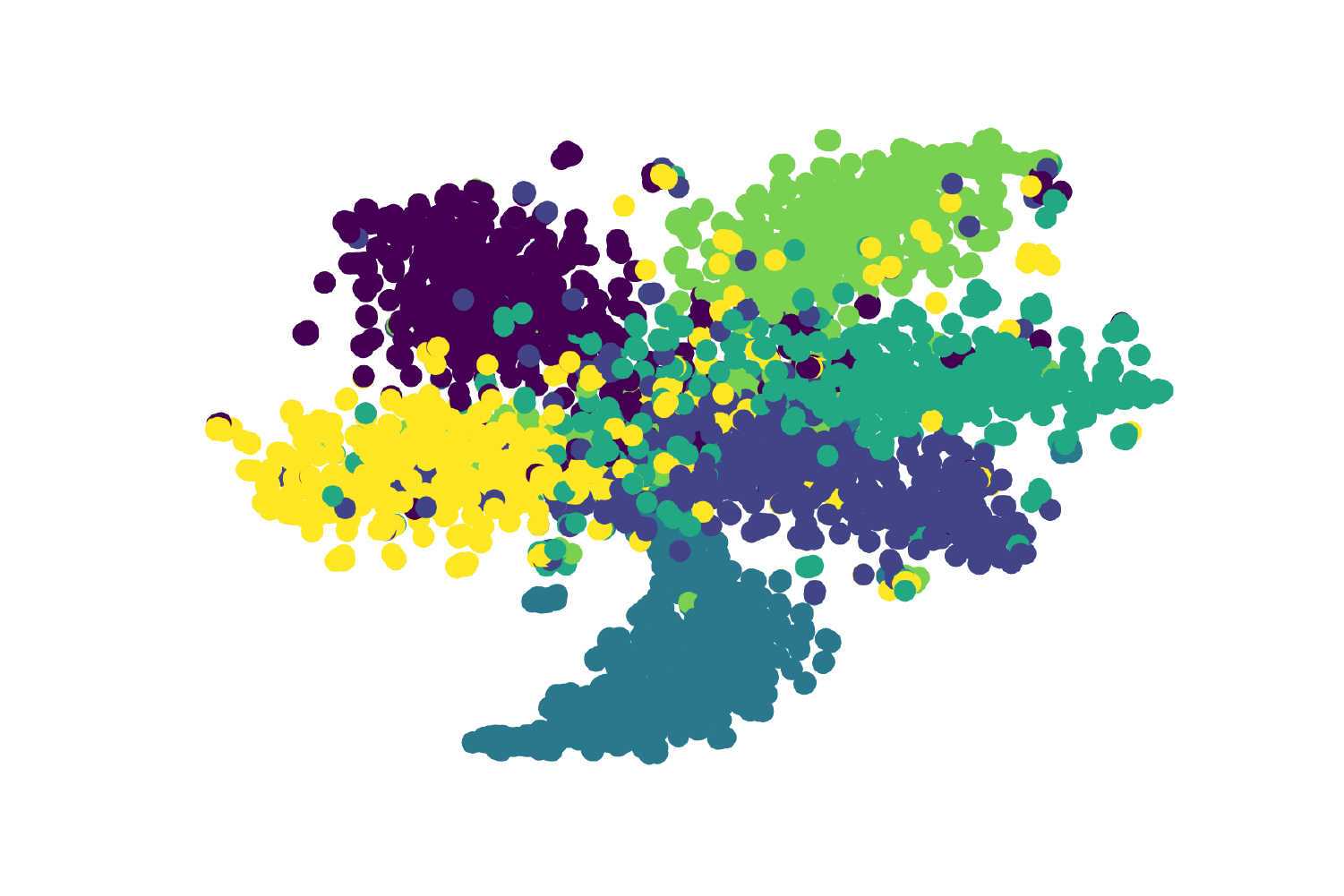}
	}
	\caption{Visualization of the learned node embeddings on BlogCatalog dataset.}
	\vspace{-5pt}
	\label{visualization_BlogCatalog}
\end{figure*}

\subsection{Node Classification}

The node classification results are reported in Table \ref{node classification}, where L/C means the number of labeled nodes per class. We have the following observations:
\begin{itemize}
\vspace{-\topsep}\item  Compared with all baselines, the proposed AM-GCN generally achieves the best performance on all datasets with all label rates. Especially, for ACC, AM-GCN achieves maximum relative improvements of 8.59\% on BlogCatalog and 8.63\% on Flickr. The results demonstrate the effectiveness of AM-GCN.
\item AM-GCN consistently outperforms GCN and \textit{k}NN-GCN on all the datasets, indicating the effectiveness of the adaptive fusion mechanism in AM-GCN, because it can extract more useful information than only performing GCN and \textit{k}NN-GCN, respectively.
\item Comparing with GCN and \textit{k}NN-GCN, we can learn that there does exist structural difference between topology graph and feature graph and performing GCN on traditional topology graph does not always show better result than on feature graph. For example, in BlogCatalog, Flickr and UAI2010, the feature graph performs better than topology. This further confirms the necessity of introducing feature graph in GCN.
\item Moreover, compared with GCN, the improvement of AM-GCN is more  substantial on the datasets with better feature graph (\textit{k}NN), such as UAI2010, BlogCatalog, Flickr. This implies that AM-GCN introduces a better and more suitable \textit{k}NN graph for label to supervise feature propagation and node representation learning.
\end{itemize}

\vspace{-\topsep}\subsection{Analysis of Variants}

In this section, we compare AM-GCN with its three variants on all datasets to validate the effectiveness of the constraints.
\begin{itemize}
\item \textbf{AM-GCN-w/o}: AM-GCN without constraints $\mathcal{L}_{c}$ and $\mathcal{L}_{d}$.
\item \textbf{AM-GCN-c}: AM-GCN with the consistency constraint $\mathcal{L}_{c}$.
\item \textbf{AM-GCN-d}: AM-GCN with the disparity constraint $\mathcal{L}_{d}$.
\end{itemize}

From the results in Figure \ref{variants}, we can draw the following conclusions: (1) The results of AM-GCN are consistently better than all the other three variants, indicating the effectiveness of using the two constraints together. (2) The results of AM-GCN-c and AM-GCN-d are usually better than AM-GCN-w/o on all datasets with all label rates, verifying the usefulness of the two constraints. (3) AM-GCN-c is generally better than AM-GCN-d on all datasets, which implies the consistency constraint plays a more vital role in this framework. (4) Comparing the results of Figure~\ref{variants} and Table~\ref{node classification}, we can find that AM-GCN-w/o, although without any constraints, still achieves very competitive performance against baselines, demonstrating that our framework is stable and competitive.

\vspace{-\topsep}\subsection{Visualization}

For a more intuitive comparison and to further show the effectiveness of our proposed model, we conduct the task of visualization on BlogCatalog dataset. We use the output embedding on the last layer of AM-GCN (or GCN, GAT) before $softmax$ and plot the learned embedding of test set using t-SNE~\cite{maaten2008visualizing}. The results of BlogCatalog in Figure \ref{visualization_BlogCatalog} are colored by real labels.

From Figure \ref{visualization_BlogCatalog}, we can find that the results of DeepWalk, GCN, and GAT are not satisfactory, because the nodes with different labels are mixed together. Apparently, the visualization of AM-GCN performs best, where the learned embedding has a more compact structure, the highest intra-class similarity and the clearest distinct boundaries among different classes.

\subsection{Analysis of Attention Mechanism}

In order to investigate whether the attention values learned by our proposed model are meaningful, we analyze the attention distribution and attention learning trend, respectively.

\begin{figure}[htbp]
	\centering
	\subfigure[Citeseer.]{
		\includegraphics[width=0.22\textwidth]{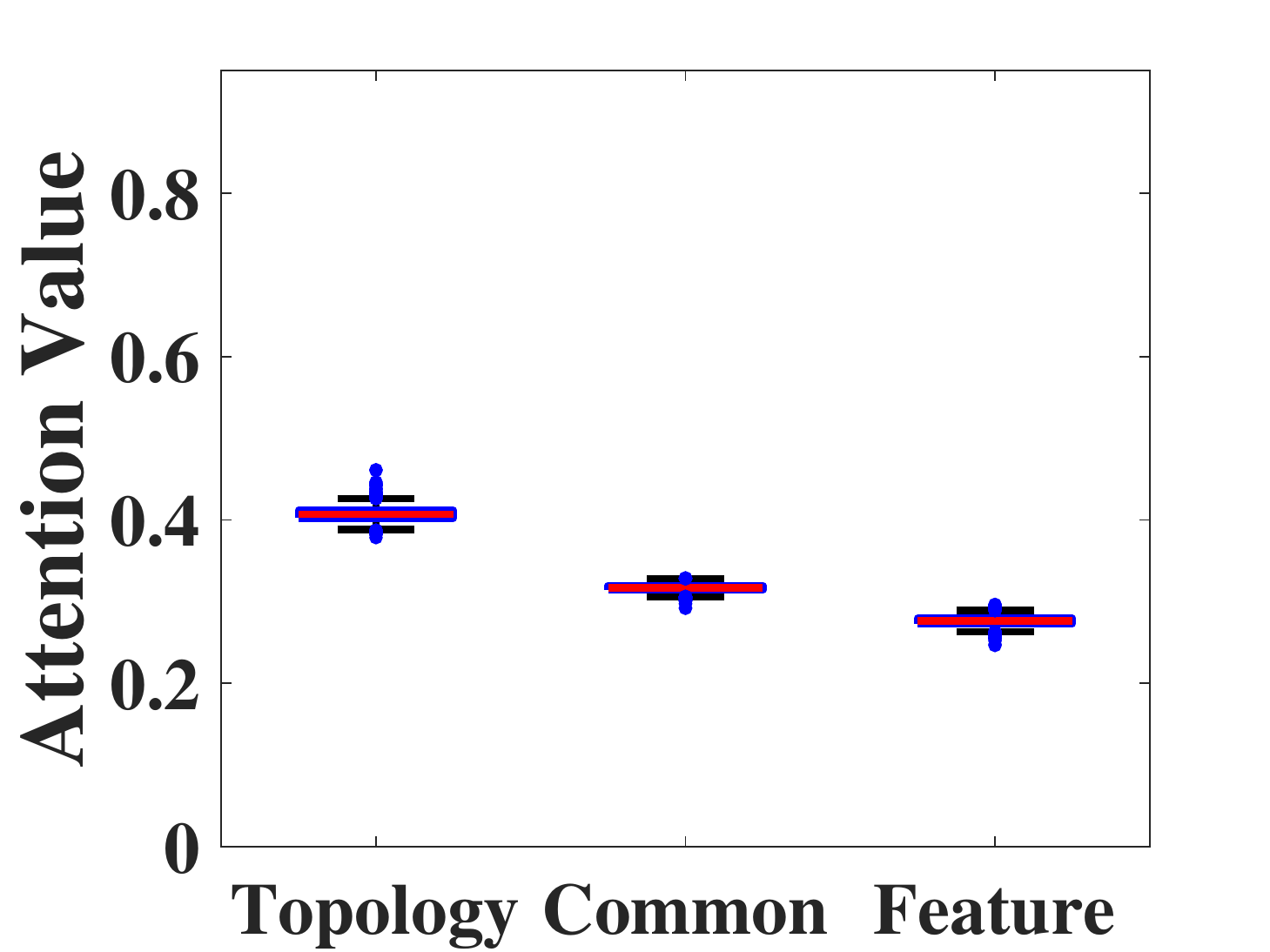}
	}
	\vspace{-5pt}
	\subfigure[UAI2010.]{
		\includegraphics[width=0.22\textwidth]{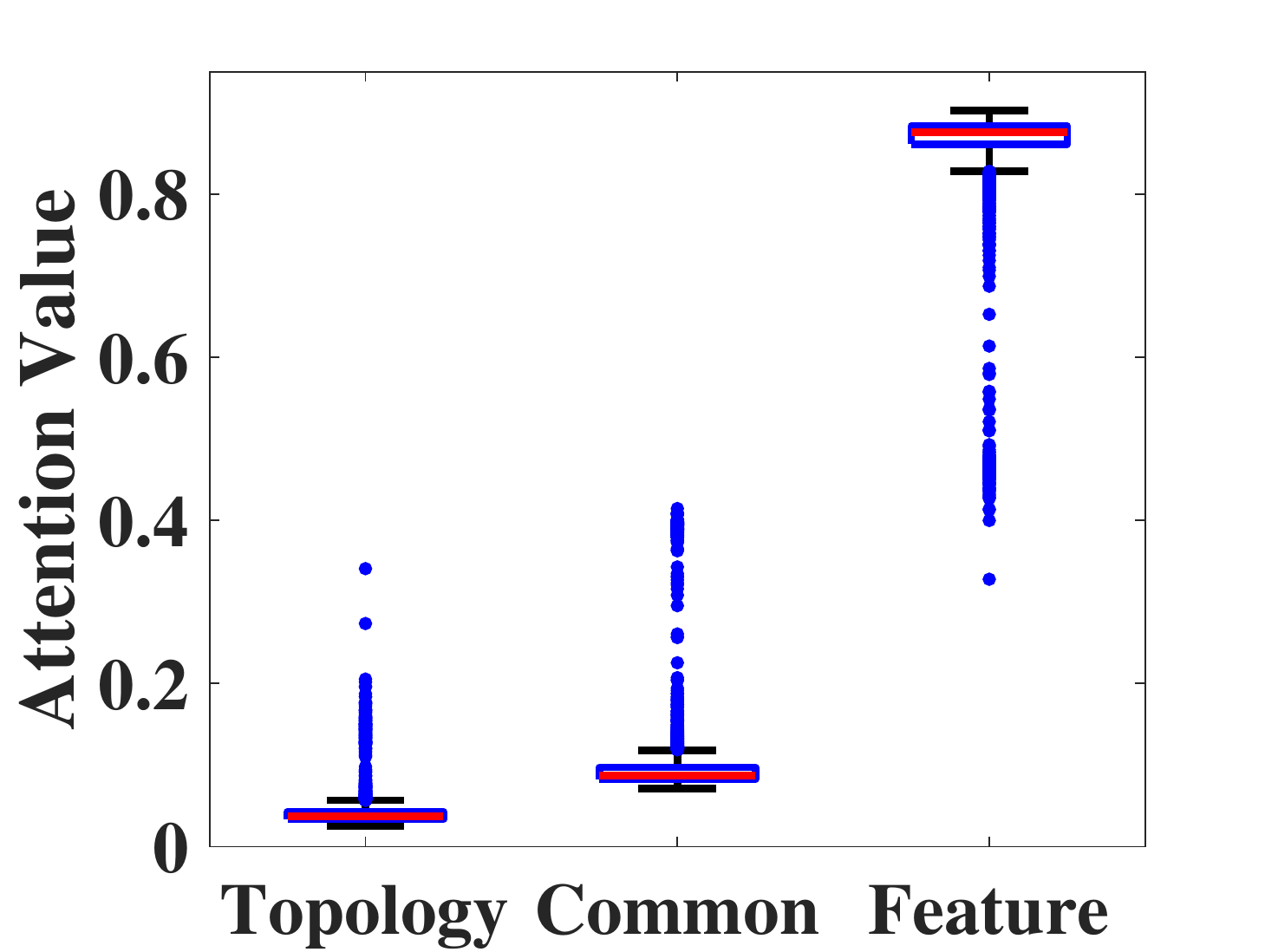}
	}
	\vspace{-5pt}
	\subfigure[ACM.]{
		\includegraphics[width=0.22\textwidth]{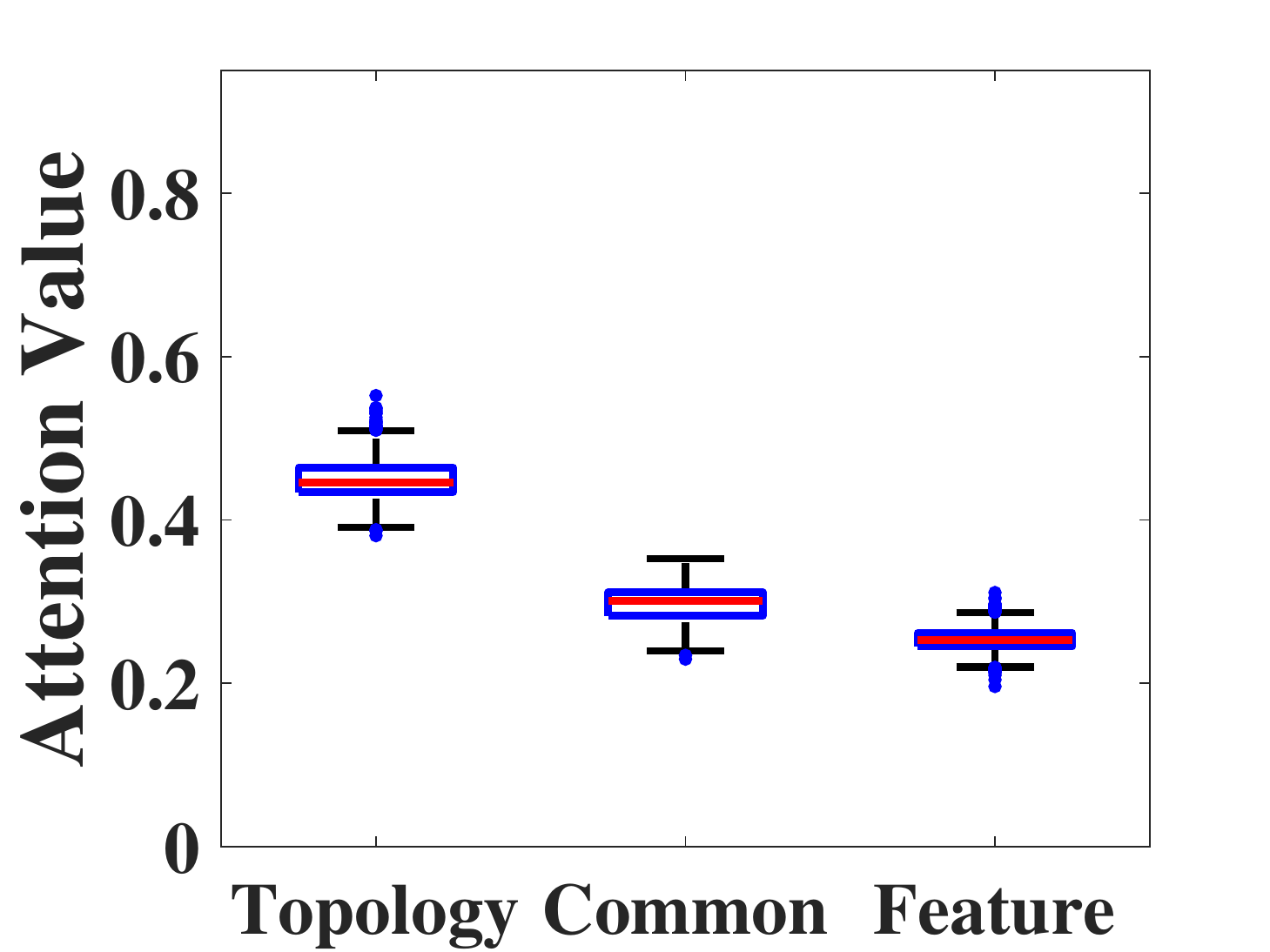}
	}
	\subfigure[BlogCatalog.]{
		\includegraphics[width=0.22\textwidth]{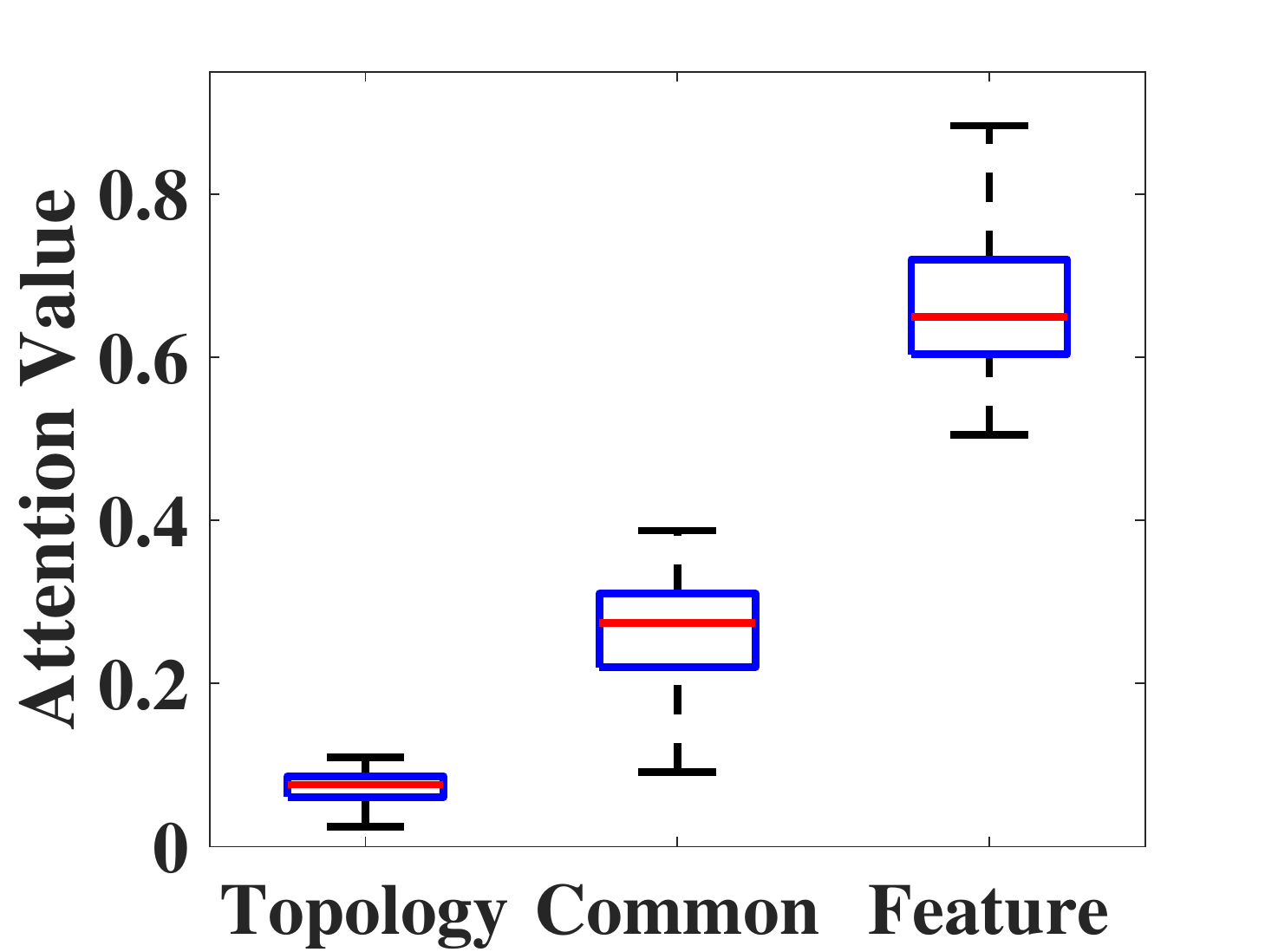}
	}
	\vspace{-5pt}
	\subfigure[Flickr.]{
		\includegraphics[width=0.22\textwidth]{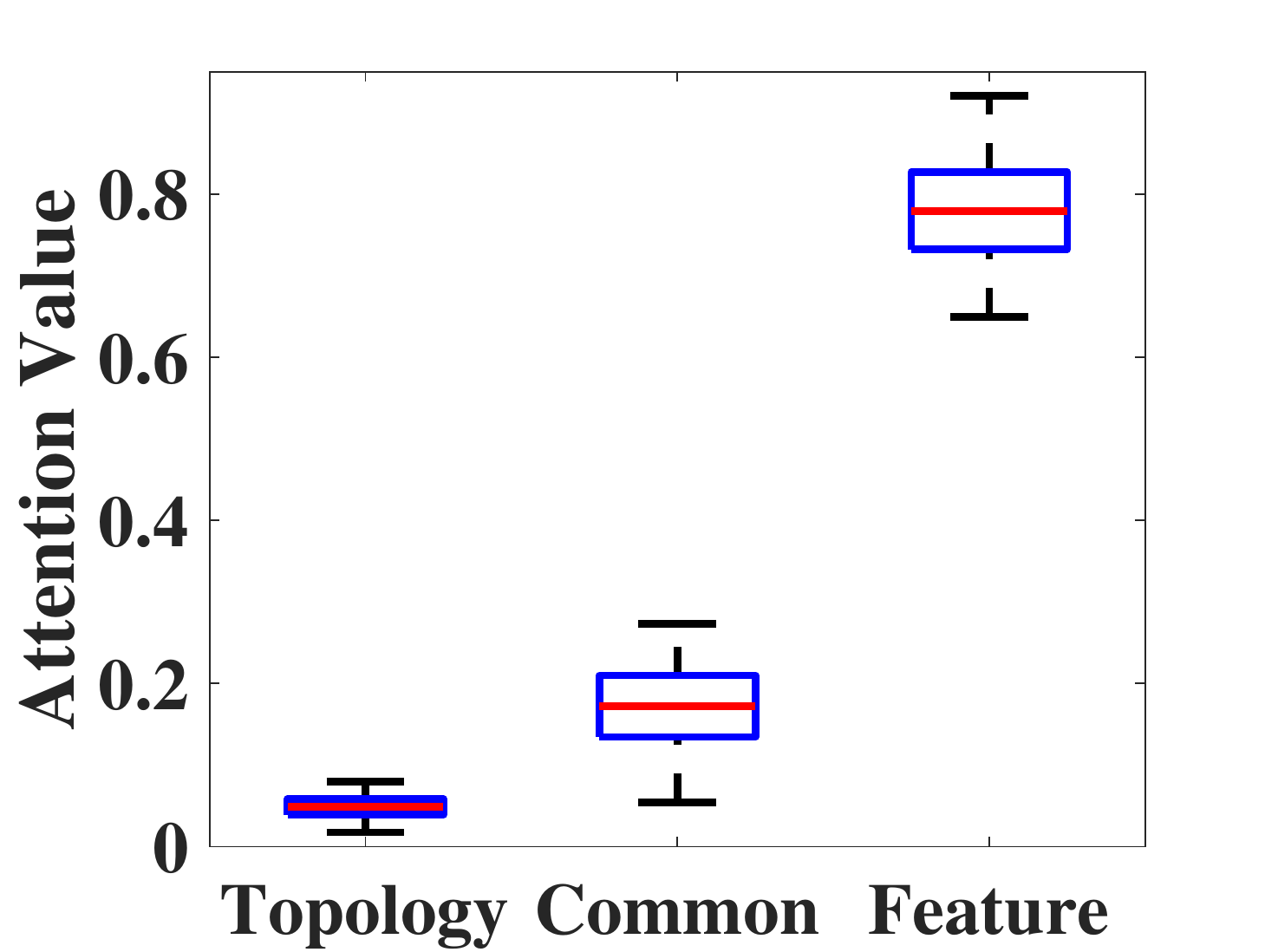}
		\label{attention_flickr}
	}
	\subfigure[CoraFull.]{
		\includegraphics[width=0.22\textwidth]{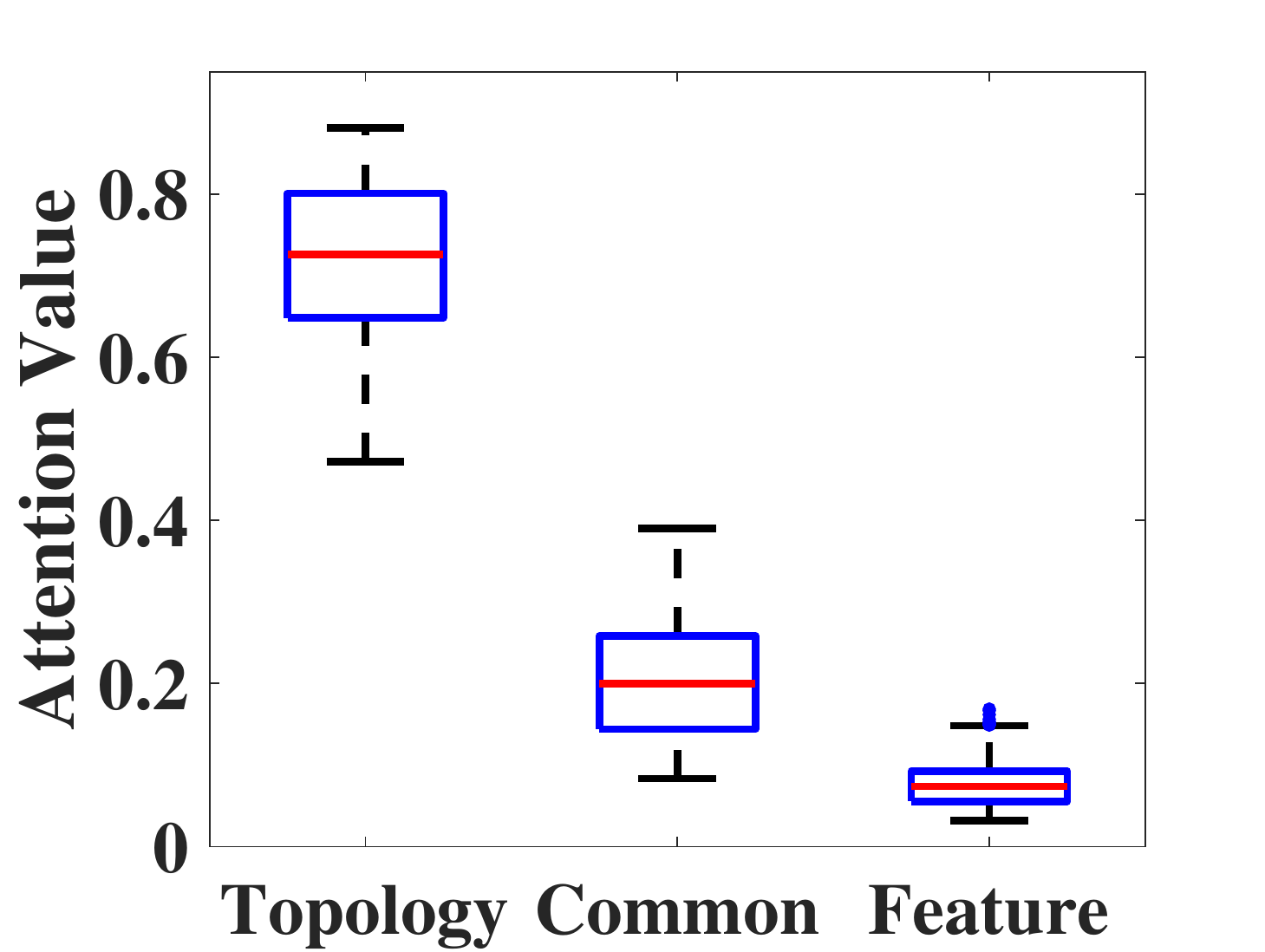}
		\label{attention_generate}
	}
	\caption{Analysis of attention distribution.}
	\vspace{-5pt}
	\label{boxplot}
\end{figure}

\begin{figure}[htbp]
	\centering
	\vspace{-5pt}
	\subfigure[\textbf{Citeseer}]{
		\includegraphics[width=0.22\textwidth]{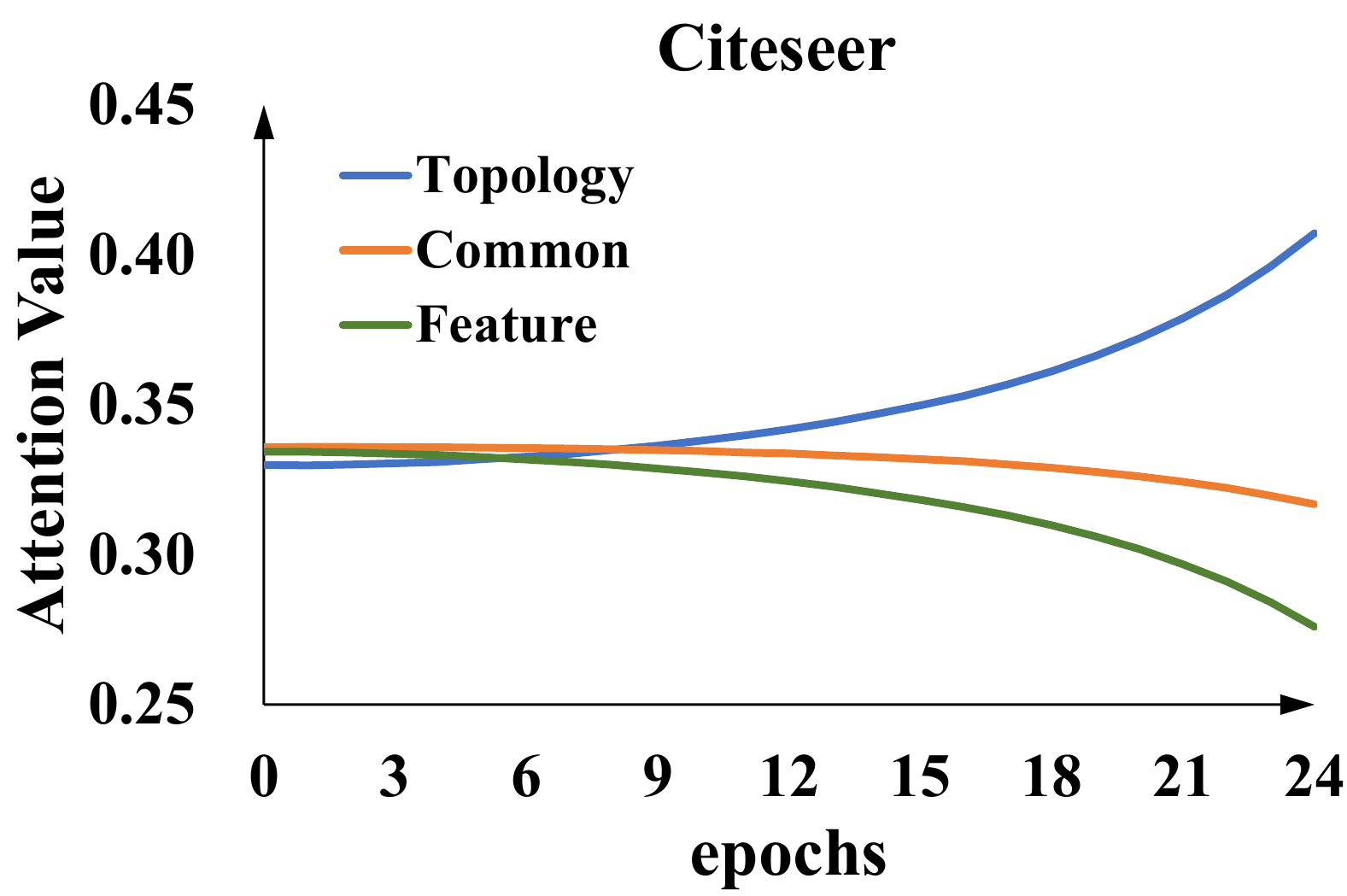}
	}
	\subfigure[\textbf{BlogCatalog}]{
		\includegraphics[width=0.22\textwidth]{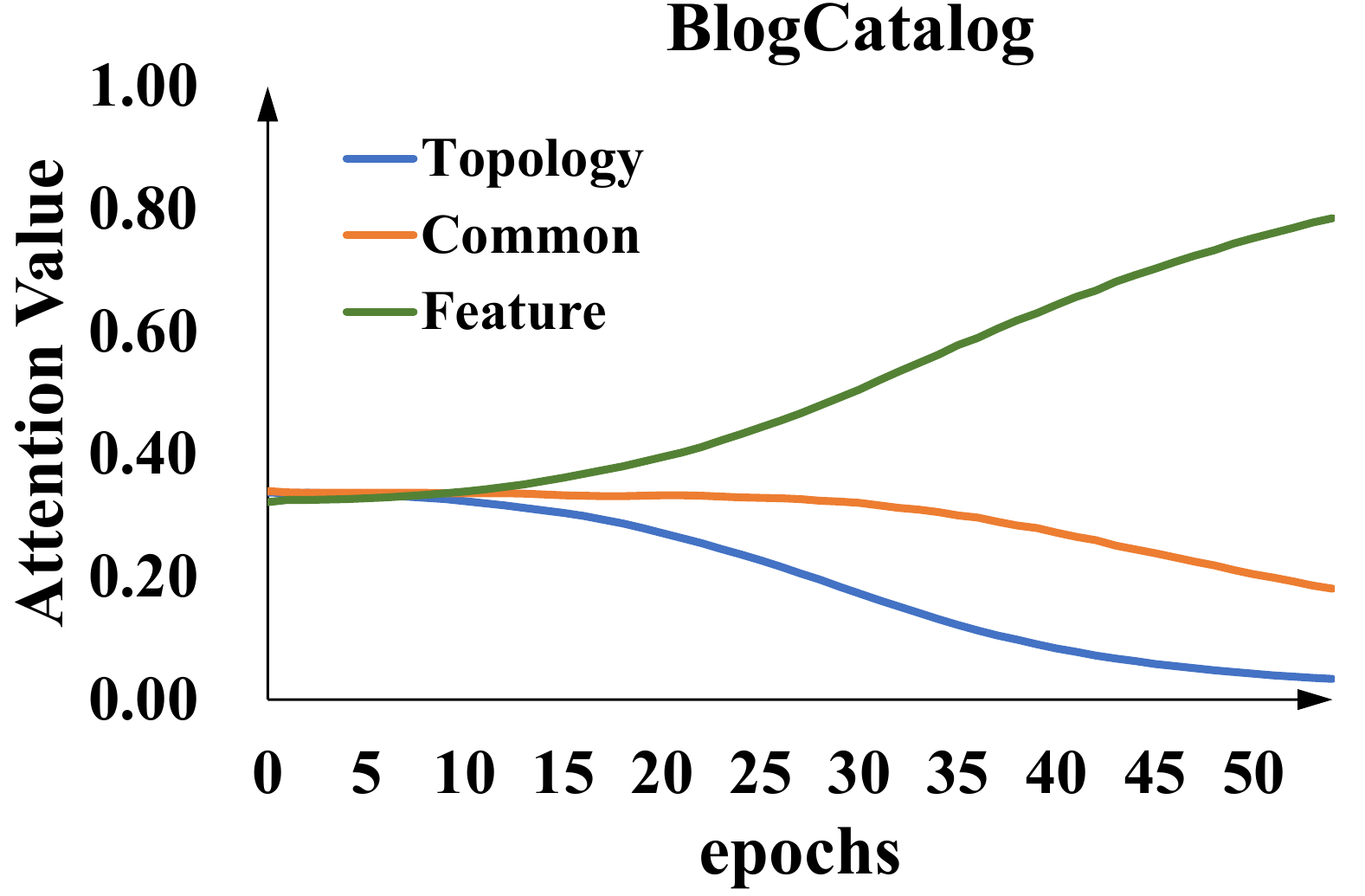}
	}
	\caption{The attention changing trends w.r.t epochs.}
	\vspace{-8pt}
	\label{attention_change}
\end{figure}

\textbf{Analysis of attention distributions.}
AM-GCN learns two specific and one common embeddings, each of which is associated with the attention values. We conduct the attention distribution analysis on all datasets with 20 label rate, where the results are shown in Figure \ref{boxplot}. As we can see, for Citeseer, ACM, CoraFull, the attention values of specific embeddings in topology space are larger than the values in feature space, and the values of common embeddings are between them. This implies that the information in topology space should be more important than the information in feature space. To verify this, we can see that the results of GCN are better than \textit{k}NN-GCN on these datasets in Table \ref{node classification}. Conversely, for UAI2010, BlogCatalog and Flickr, in comparison with Figure \ref{boxplot} and Table \ref{node classification}, we can find \textit{k}NN-GCN performs better than GCN, meanwhile, the attention values of specific embeddings in feature space are also larger than those in topology space. In summary, the experiment demonstrates that our proposed AM-GCN is able to adaptively assign larger attention value for more important information.

\textbf{Analysis of attention trends.}
We analyze the changing trends of attention values during the training process. Here we take Citeseer and BlogCatalog with 20 label rate as examples in Figure \ref{attention_change}, where \textbf{x}-axis is the epoch and \textbf{y}-axis is the average attention value. More results are in supplement~\ref{trends}. At the beginning, the average attention values of Topology, Feature, and Common are almost the same, with the training epoch increasing, the attention values become different. For example, in BlogCatalog, the attention value for topology gradually decreases, while the attention value for feature keeps increasing. This phenomenon is consistent with the conclusions in Table \ref{node classification} and Figure \ref{boxplot}, i.e., \textit{k}NN-GCN with feature graph performs better than GCN and the information in feature space is more important than in topology space. We can see that AM-GCN can learn the importance of different embeddings step by step.

\begin{figure}[t]
	\centering
	\vspace{-5pt}
	\subfigure[\textbf{Citeseer}]{
		\includegraphics[width=0.22\textwidth]{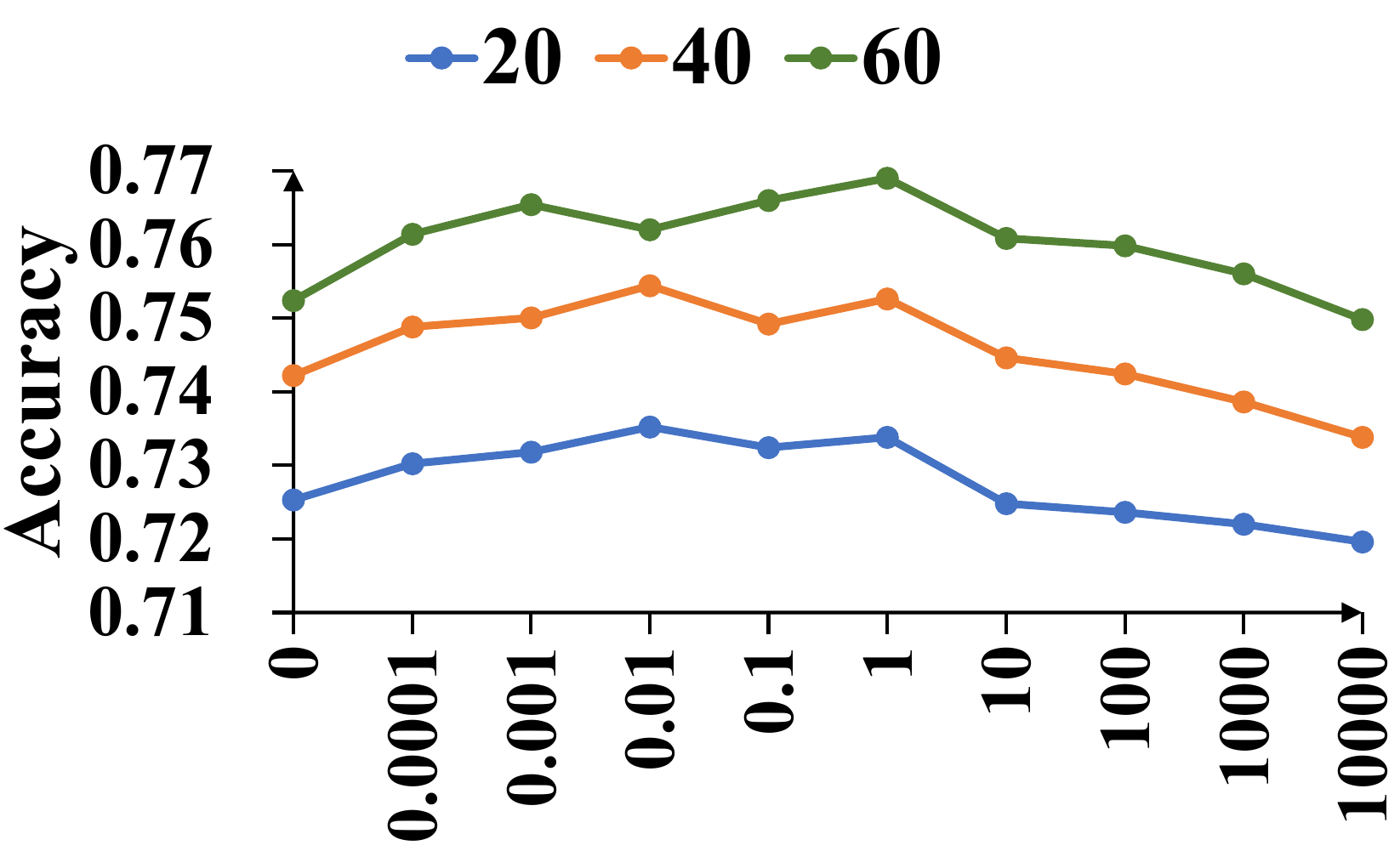}
	}
	\subfigure[\textbf{BlogCatalog}]{
		\includegraphics[width=0.22\textwidth]{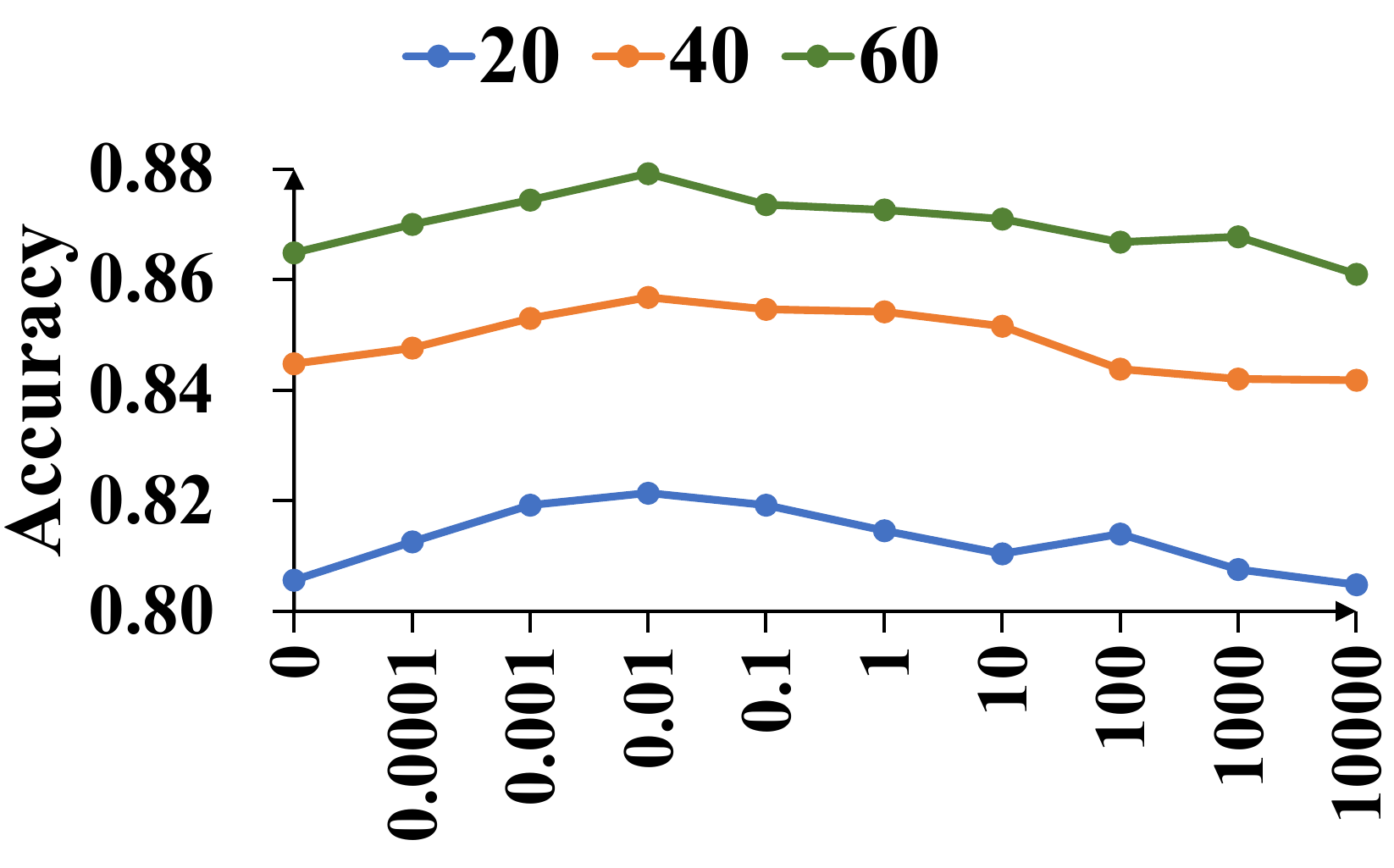}
	}
	\caption{Analysis of parameter \textit{$\gamma$}.}
	\vspace{-5pt}
	\label{alpha}
\end{figure}

\begin{figure}[t]
	\centering
	\vspace{-5pt}
	\subfigure[\textbf{Citeseer}]{
		\includegraphics[width=0.22\textwidth]{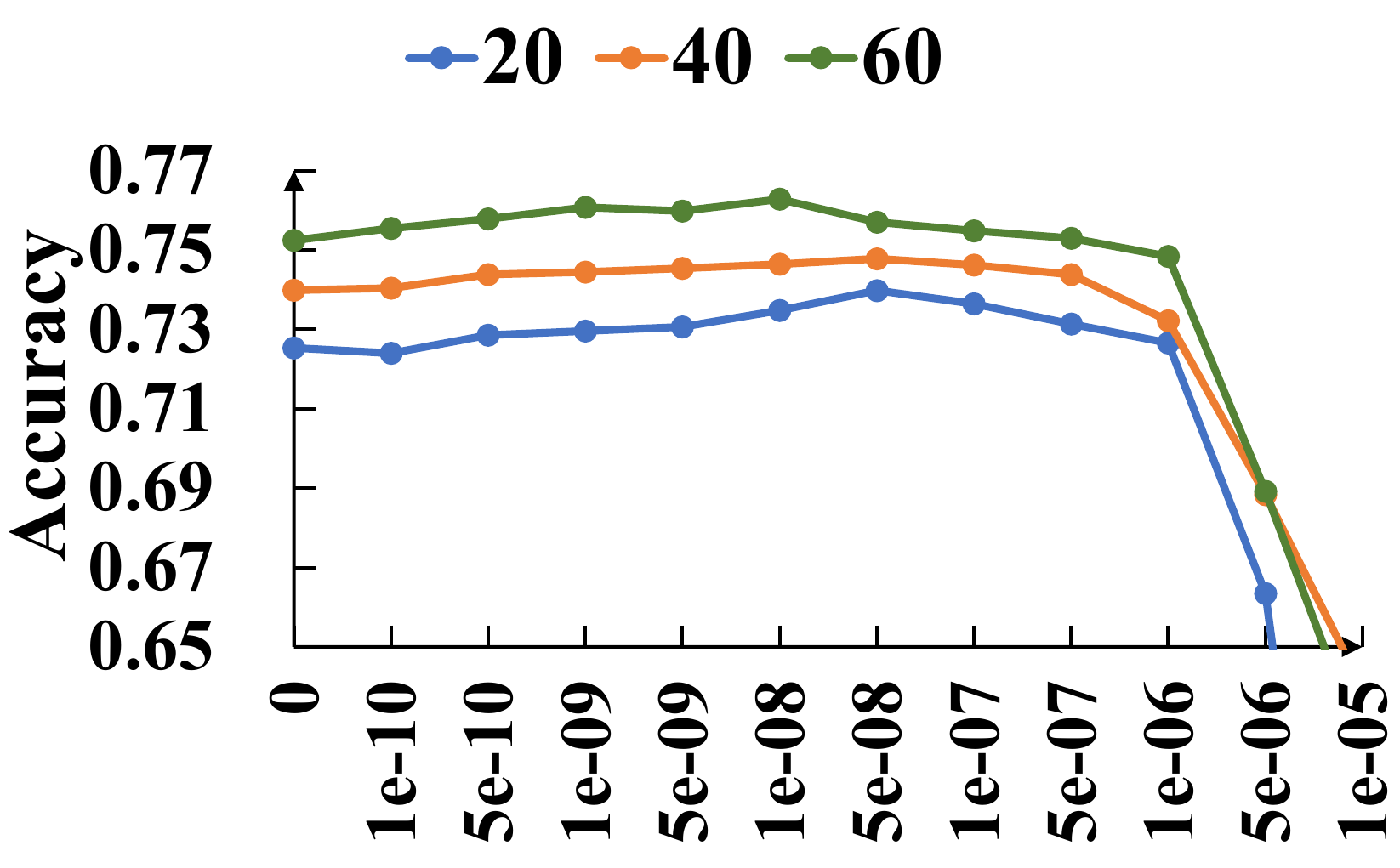}
		\label{citeseer}
	}
	\subfigure[\textbf{BlogCatalog}]{
		\includegraphics[width=0.22\textwidth]{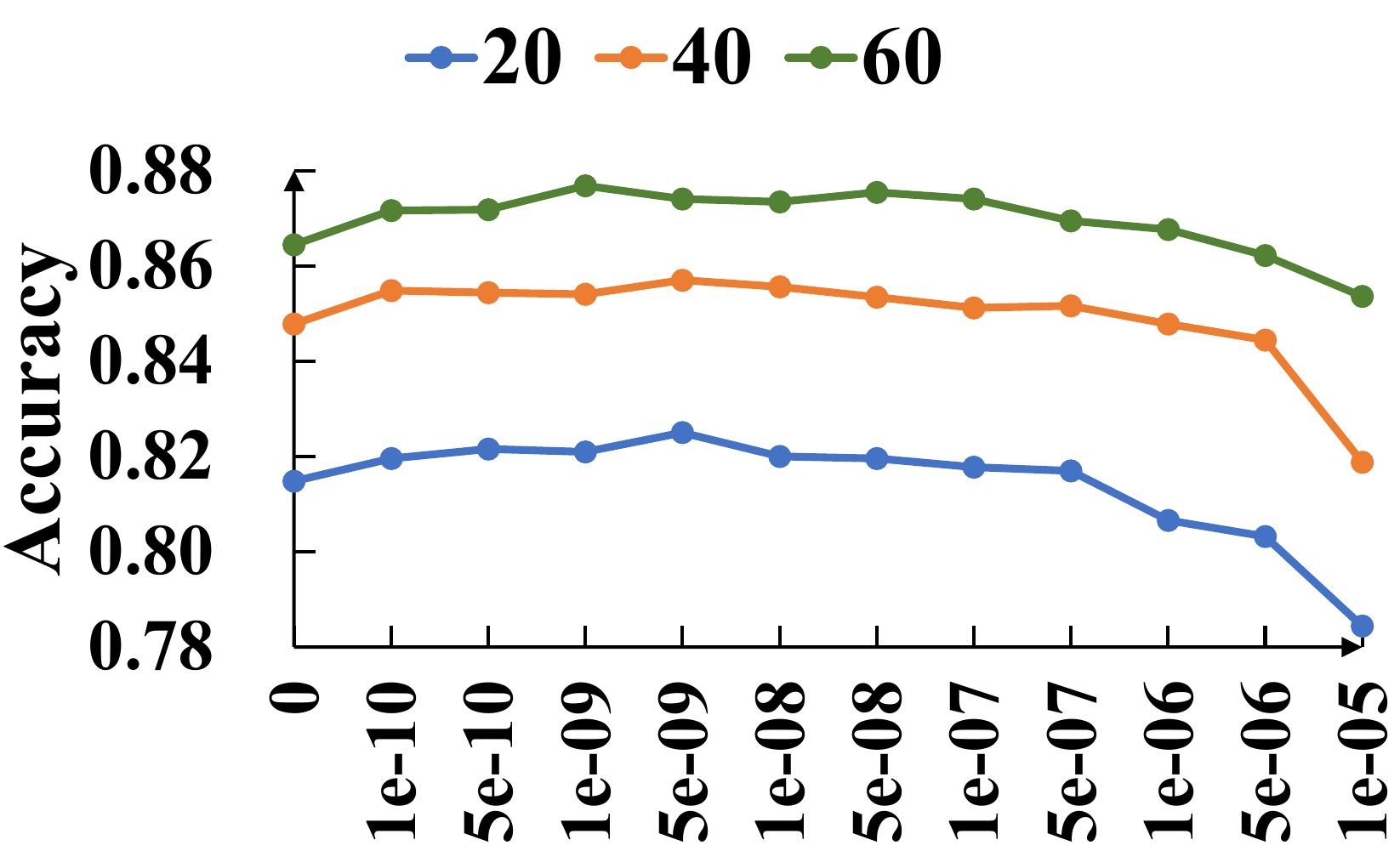}
	}
	\caption{Analysis of parameter \textit{$\beta$}.}
	\vspace{-5pt}
	\label{beta}
\end{figure}

\begin{figure}[t]
	\centering
	\vspace{-5pt}
	\subfigure[\textbf{Citeseer}]{
		\includegraphics[width=0.22\textwidth]{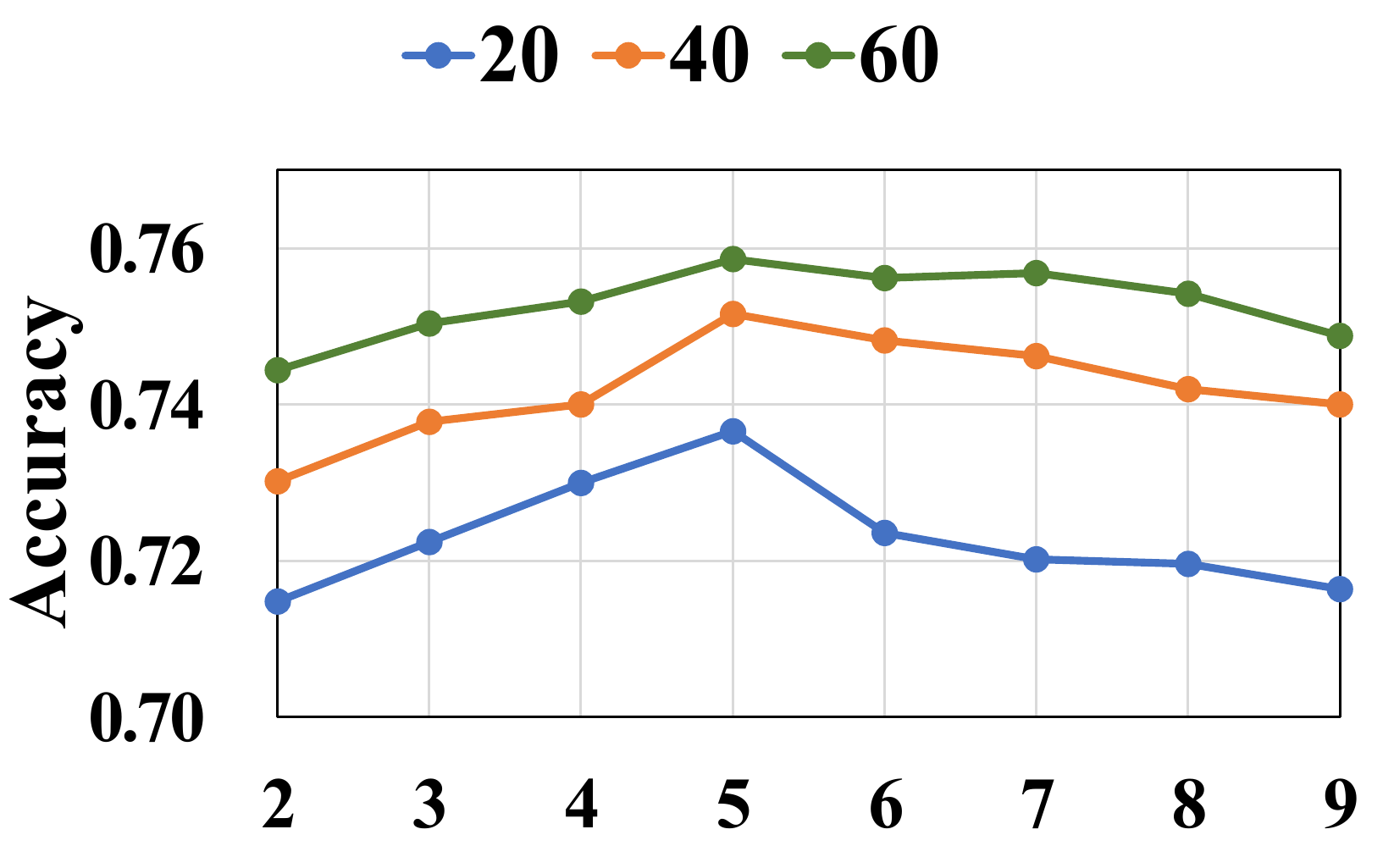}
		%\caption{fig1}
	}
	\subfigure[\textbf{BlogCatalog}]{
		\includegraphics[width=0.22\textwidth]{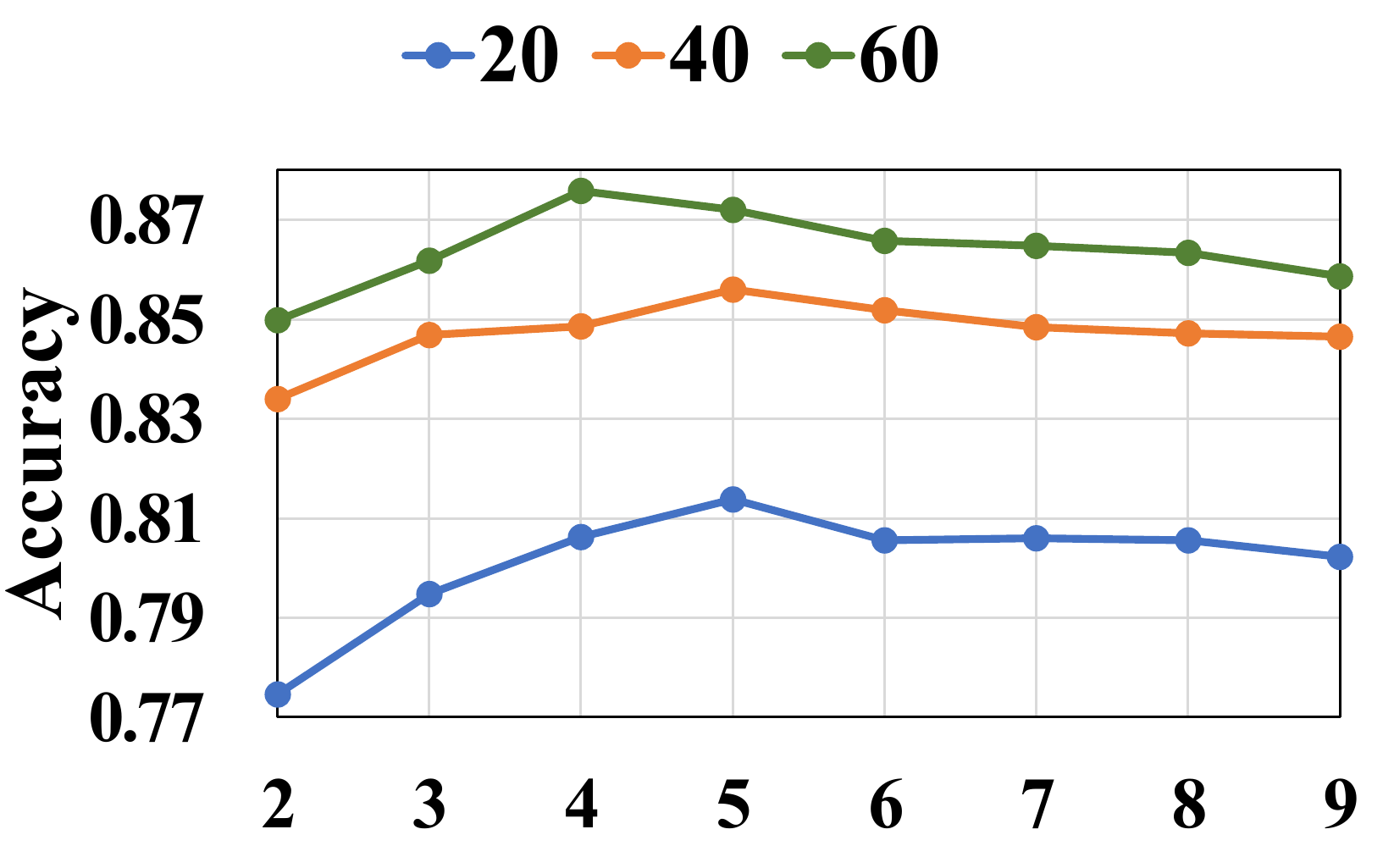}
	}
	\caption{Analysis of parameter \textit{k}.}
	\vspace{-8pt}
	\label{k}
\end{figure}

\vspace{-\topsep}\subsection{Parameter Study}\label{sec:parameter study}
In this section, we investigate the sensitivity of parameters on Citeseer and BlogCatalog datasets. More results are in \ref{ps}.

\textbf{Analysis of consistency coefficient} \textit{$\gamma$}.
We test the effect of the consistency constraint weight $\gamma$ in Eq. \eqref{final_target}, and vary it from 0 to 10000. The results are shown in Figure \ref{alpha}. With the increase of the consistency coefficient, the performances raise first and then start to drop slowly. Basically, AM-GCN is stable when the \textit{$\gamma$} is within the range from 1e-4 to 1e+4 on all datasets. We can also see that the curves of 20, 40, 60 label rates show similar changing trend.

\textbf{Analysis of disparity constraint coefficient} \textit{$\beta$}.
We then test the effect of the disparity constraint weight $\beta$ in Eq. \eqref{final_target}, and vary it from 0 to 1e-5. The results are shown in Figure \ref{beta}. Similarly, with the increase of \textit{$\beta$}, the performances also raise first, but the performance will drop quickly if \textit{$\beta$} is larger than 1e-6 for Citeseer in Figure \ref{citeseer}, while for BlogCatalog, it is relatively stable.

\textbf{Analysis of \textit{k}-nearest neighbor graph} \textit{k}.
In order to check the impact of the top \textit{k} neighborhoods in \textit{k}NN graph, we study the performance of AM-GCN with various number of \textit{k} ranging from 2 to 10 in Figure \ref{k}. For Citeseer and BlogCatalog, the accuracies increase first and then start to decrease. It may probably because that if the graph becomes denser, the feature is easier to be smoothed, and also, larger \textit{k} may introduce more noisy edges.

\section{Related Work}\label{sec:related-work}

Recently, graph convolutional network (GCN) models~\cite{gao2018large, chen2018fastgcn, ma2019graph, qu2019gmnn, xu2019how, ying2019hyperbolic} have been widely studied. For example, \cite{bruna2014spectral} first designs the graph convolution operation in Fourier domain by the graph Laplacian. Then \cite{defferrard2016convolutional} further employs the Chebyshev expansion of the graph Laplacian to improve the efficiency. \cite{kipf2017semi} simplifies the convolution operation and proposes to only aggregate the node features from the one-hop neighbors. GAT \cite{ve2018graph} introduces the attention mechanism to aggregate node features with the learned weights. GraphSAGE \cite{hamilton2017inductive} proposes to sample and aggregate features from local neighborhoods of nodes with mean/max/LSTM pooling. DEMO-Net \cite{wu2019demo} designs a degree-aware feature aggregation process. MixHop \cite{abu-el-haija2019mixhop} aggregates feature information from both first-order and higher-order neighbors in each layer of network, simultaneously. Most of the current GCNs essentially focus on fusing network topology and node features to learn node embedding for classification. Also, there are some recent works on analyzing the fusion mechanism of GCN. For example, ~\cite{li2018deeper} shows that GCNs actually perform the Laplacian smoothing on node features, ~\cite{nt2019revisiting} and ~\cite{wu2019simplifying} prove that topological structures play the role of low-pass filtering on node features. To learn more works on GCNs, please refer to the elaborate reviews \cite{zhang2018deep,wu2019comprehensive}. However, whether the GCNs can adaptively extract the correlated information from node features and topological structures for classification remains unclear.

\section{Conclusion}\label{sec:con}
\balance
In this paper, we rethink the fusion mechanism of network topology and node features in GCN and surprisingly discover it is distant from optimal. Motivated by this fundamental problem, we study how to adaptively learn the most correlated information from topology and node features and sufficiently fuse them for classification. We propose a multi-channel model AM-GCN which is able to learn suitable importance weights when fusing topology and node feature information. Extensive experiments well demonstrate the superior performance over the state-of-the-art models on real world datasets.

\section{Acknowledgments}
This work is supported in part by the National Natural Science Foundation of China (No. 61702296, 61772082, 61806020, U1936104, U1936219, 61772304, U1611461, 61972442), the National Key Research and Development Program of China (No. 2018YFB1402600, 2018AAA0102004), the CCF-Tencent Open Fund, Beijing Academy of Artificial Intelligence (BAAI), and a grant from the Institute for Guo Qiang, Tsinghua University. Jian Pei's research is supported in part by the NSERC Discovery Grant program. All opinions, findings, conclusions and recommendations are those of the authors and do not necessarily reflect the views of the funding agencies.

%%
%% The acknowledgments section is defined using the "acks" environment
%% (and NOT an unnumbered section). This ensures the proper
%% identification of the section in the article metadata, and the
%% consistent spelling of the heading.

%%
%% The next two lines define the bibliography style to be used, and
%% the bibliography file.

\bibliographystyle{ACM-Reference-Format}
\bibliography{amgcn}

%%
%% If your work has an appendix, this is the place to put it.

\newpage
\appendix
\section{SUPPLEMENT}

In the supplement, for the reproducibly, we provide our experimental environment and all the baselines and datasets websites. The implementation details, including the detailed hyper-parameter values for all the experiments, are also provided. Finally, we show more additional results to support the conclusions in our paper.

\subsection{Experiments Settings}
All experiments are conducted with the following setting:
\begin{itemize}
	\item Operating system: CentOS Linux release 7.6.1810
	\item CPU: Intel(R) Xeon(R) CPU E5-2620 v4 @ 2.10GHz
	\item GPU: GeForce GTX 1080 Ti
	\item Software versions: Python 3.7; Pytorch 1.1.0; Numpy 1.16.2; SciPy 1.3.1; NetworkX 2.4; scikit-learn 0.21.3
\end{itemize}

\subsection{Baselines and Datasets}

The publicly available implementations of Baselines can be found at the following URLs:
\begin{itemize}
	\item DeepWalk, LINE: \url{https://github.com/thunlp/OpenNE}
	\item Chebyshev: \url{https://github.com/tkipf/gcn}
	\item GCN in Pytorch: \url{https://github.com/tkipf/pygcn}
	\item GAT in Pytorch: \url{https://github.com/Diego999/pyGAT/}
	\item DEMO-Net: \url{https://github.com/jwu4sml/DEMO-Net}
	\item MixHop: \url{https://github.com/samihaija/mixhop}
\end{itemize}
And the datasets used in this paper can be found as the following URLs:
\begin{itemize}
	\item Citeseer: \url{https://github.com/tkipf/pygcn}
	\item UAI2010: \url{http://linqs.umiacs.umd.edu/projects//projects/lbc/index.html}
	\item ACM: \url{https://github.com/Jhy1993/HAN}
	\item BlogCatalog: \url{https://github.com/mengzaiqiao/CAN}
	\item Flickr: \url{https://github.com/mengzaiqiao/CAN}
	\item CoraFull: \url{https://github.com/abojchevski/graph2gauss/}
\end{itemize}

\subsection{Implementation Details} \label{sec:parameters}
The codes of AM-GCN are based on the
Graph Convolutional Networks in PyTorch version\footnote{https://github.com/tkipf/pygcn}. And for the reproducibility of our proposed model, we also list the parameter values used in our model in Tabel \ref{hyperparameters}.

\subsection{Additional Results}
In this section, we provide the additional results of our experiments including analysis of attention trends on the other four datasets and parameter study on UAI2010 and Flickr datasets.

\begin{figure}[b]
	\vspace{2pt}
	\centering
	\subfigure[\textbf{UAI2010}]{
		\includegraphics[width=0.22\textwidth]{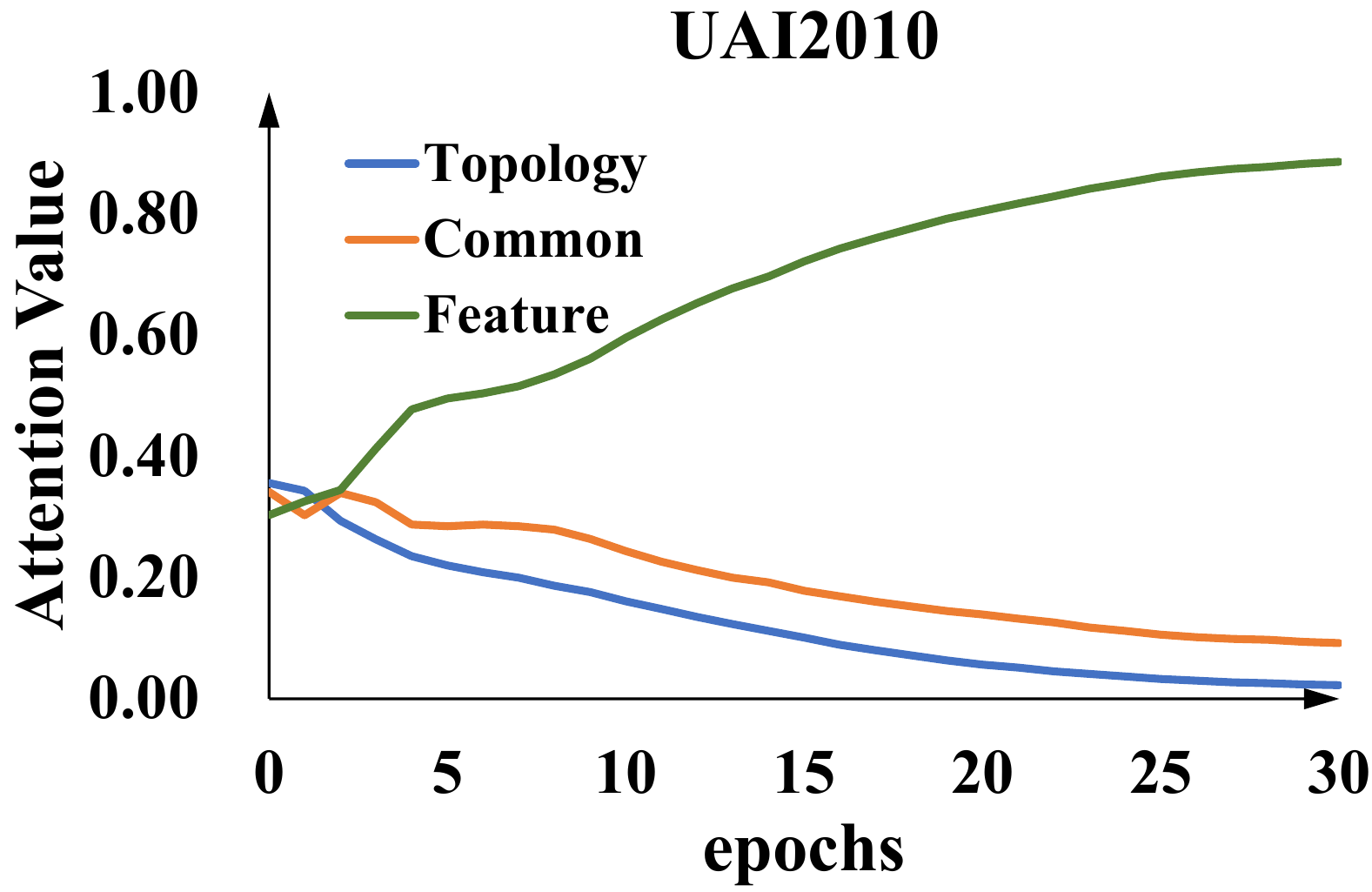}
	}
	\subfigure[\textbf{ACM}]{
		\includegraphics[width=0.22\textwidth]{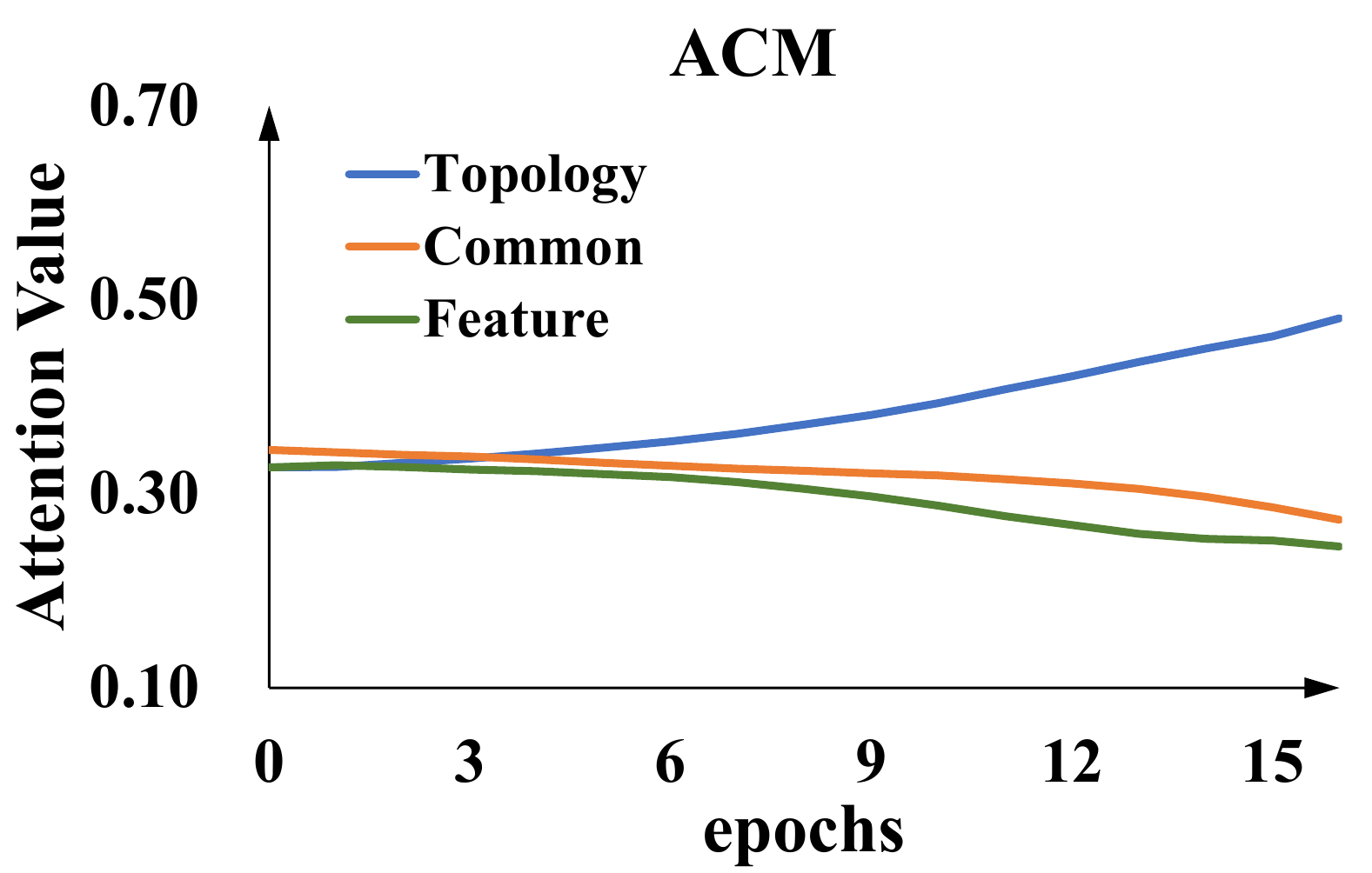}
	}
	\quad
	\vspace{2pt}
	\subfigure[\textbf{Flickr}]{
		\includegraphics[width=0.22\textwidth]{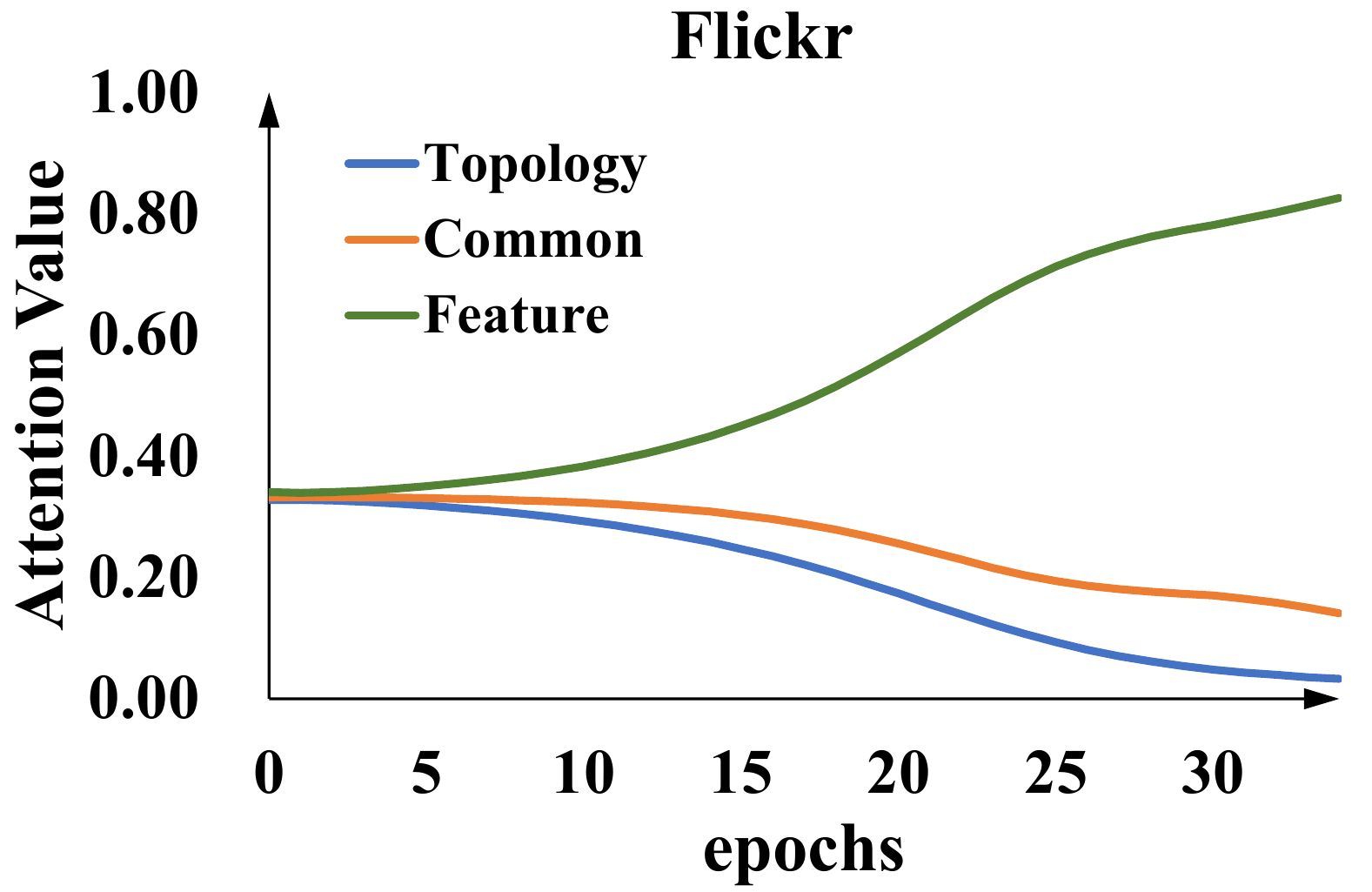}
	}
	\subfigure[\textbf{CoraFull}]{
		\includegraphics[width=0.22\textwidth]{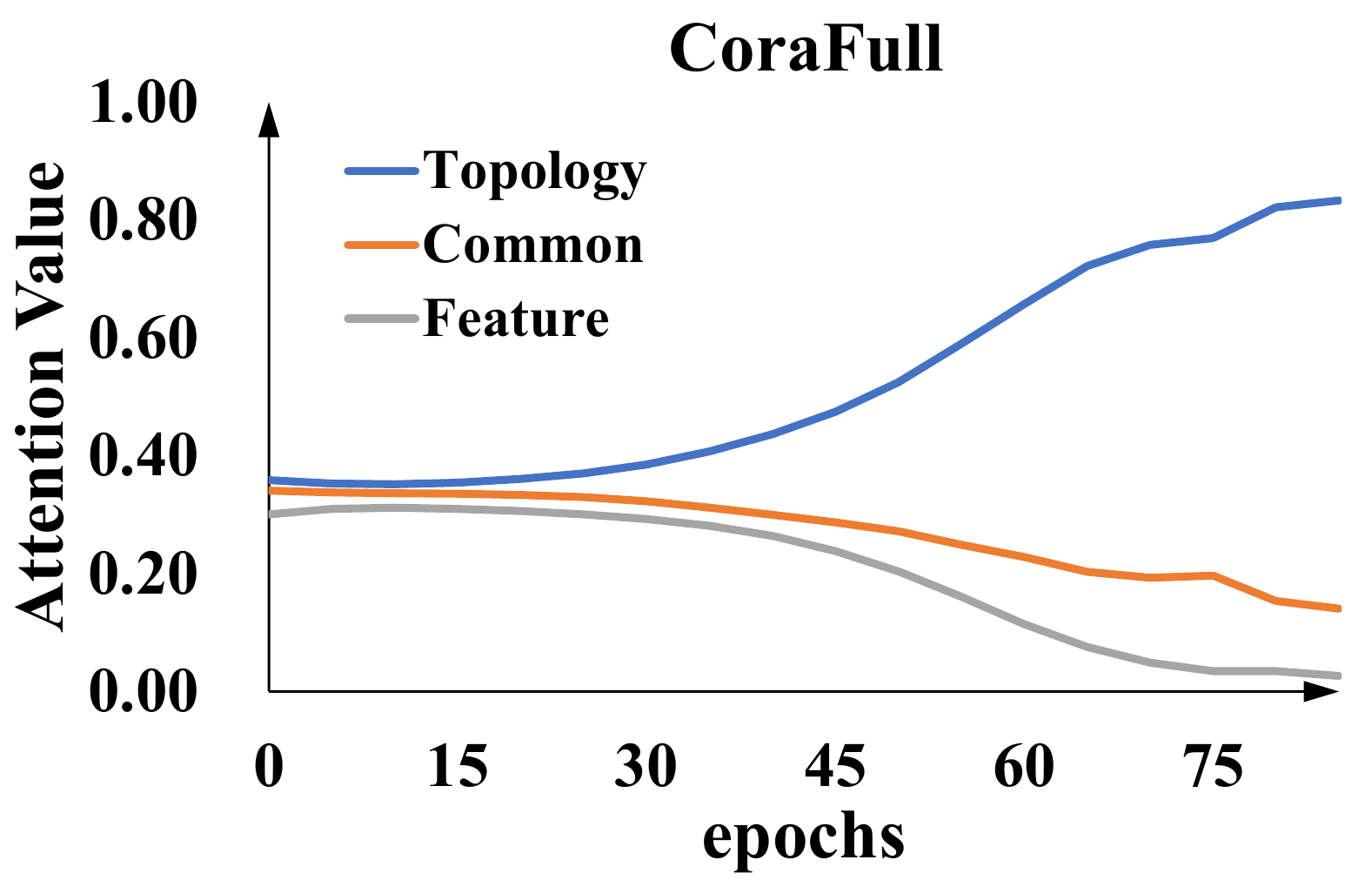}
	}
	\caption{Changing trends on another four datasets.}
	\vspace{2pt}
	\label{attention_change_2}
\end{figure}

\begin{figure}[b]
	\centering
	\vspace{2pt}
	\subfigure[\textbf{UAI2010}]{
		\includegraphics[width=0.22\textwidth]{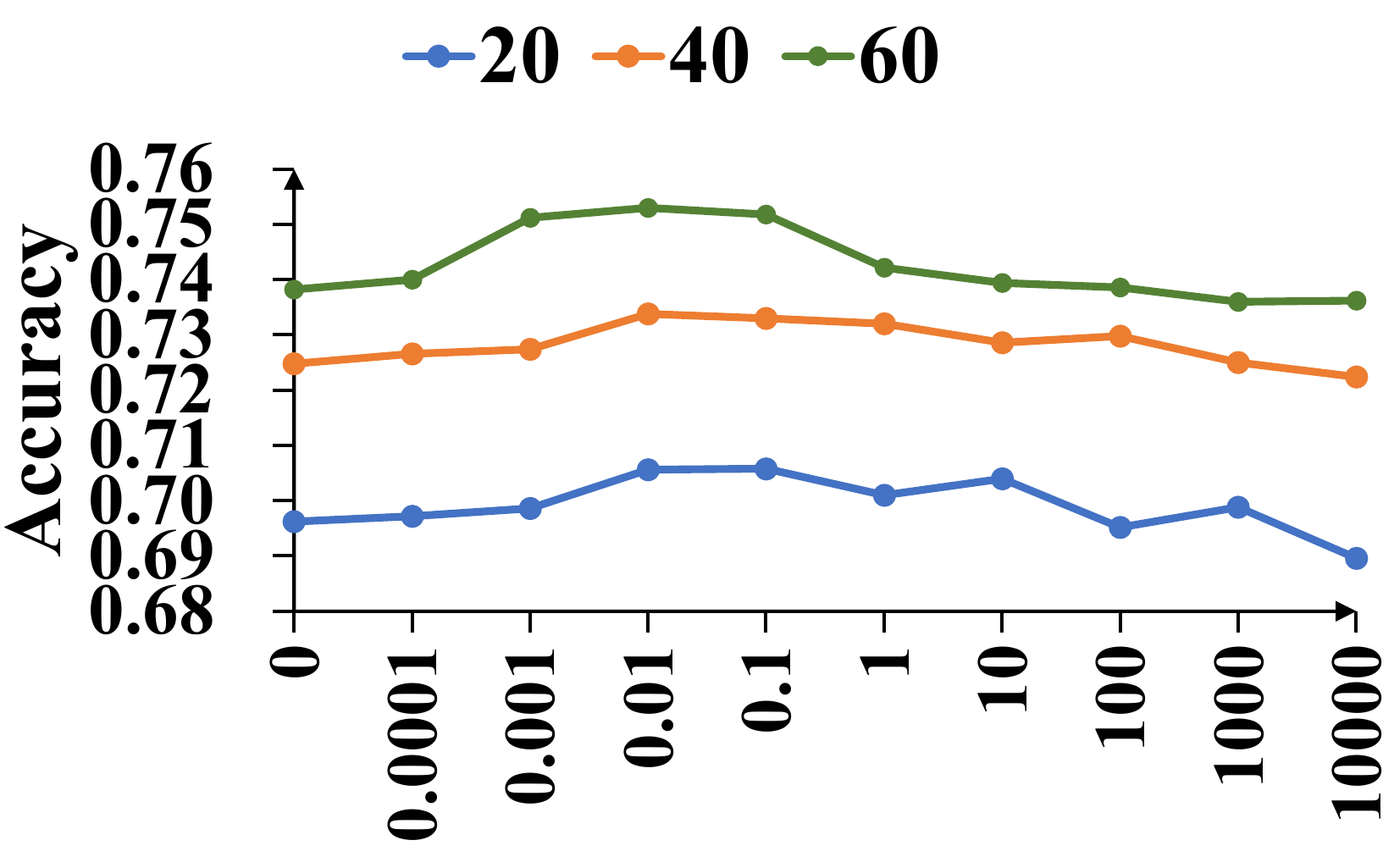}
	}
	\subfigure[\textbf{Flickr}]{
		\includegraphics[width=0.22\textwidth]{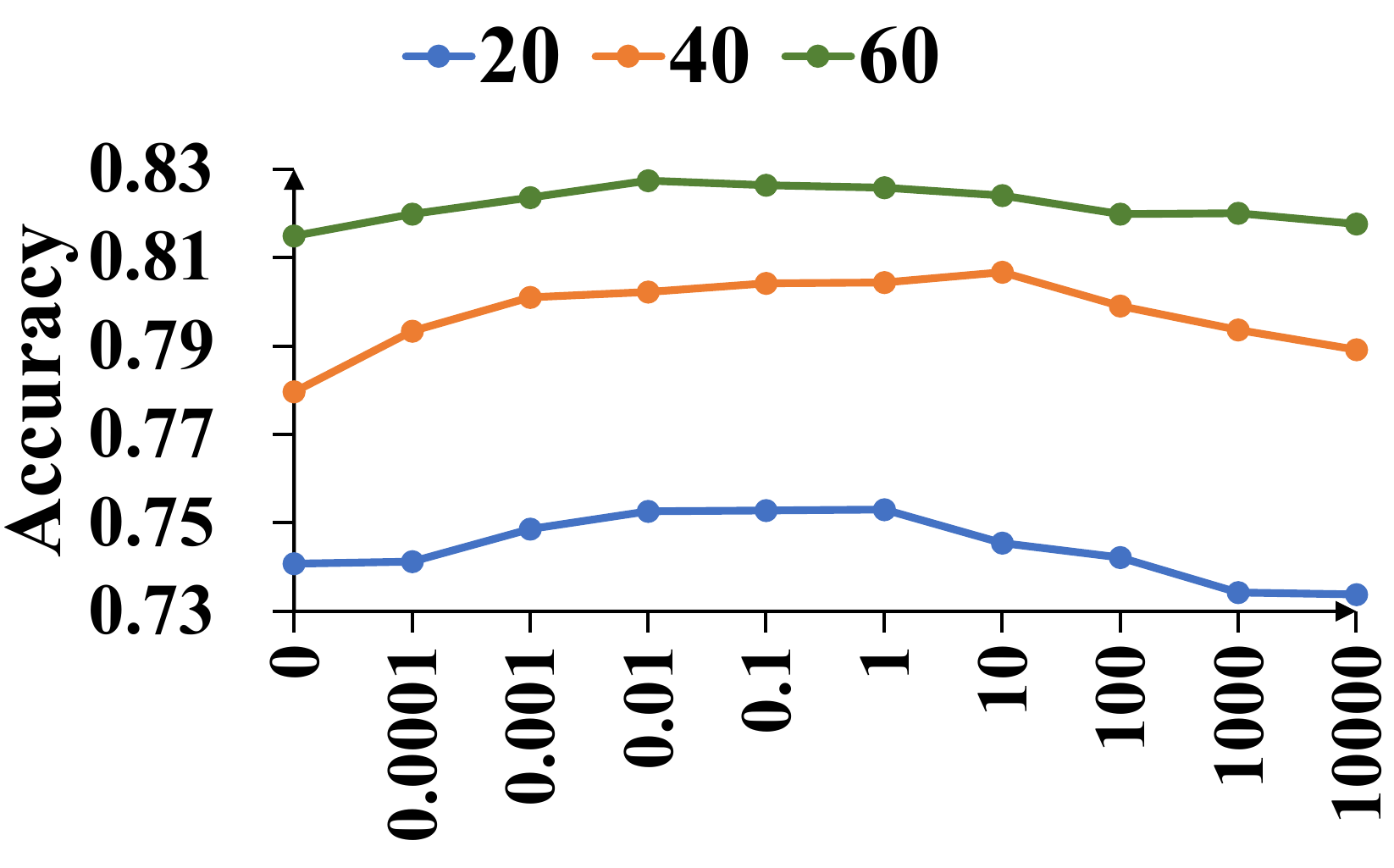}
	}
	\caption{Analysis of parameter \textit{$\gamma$}.}
	\vspace{2pt}
	\label{alpha_2}
\end{figure}
\begin{figure}[b]
	\centering
	\vspace{2pt}
	\subfigure[\textbf{UAI2010}]{
		\includegraphics[width=0.22\textwidth]{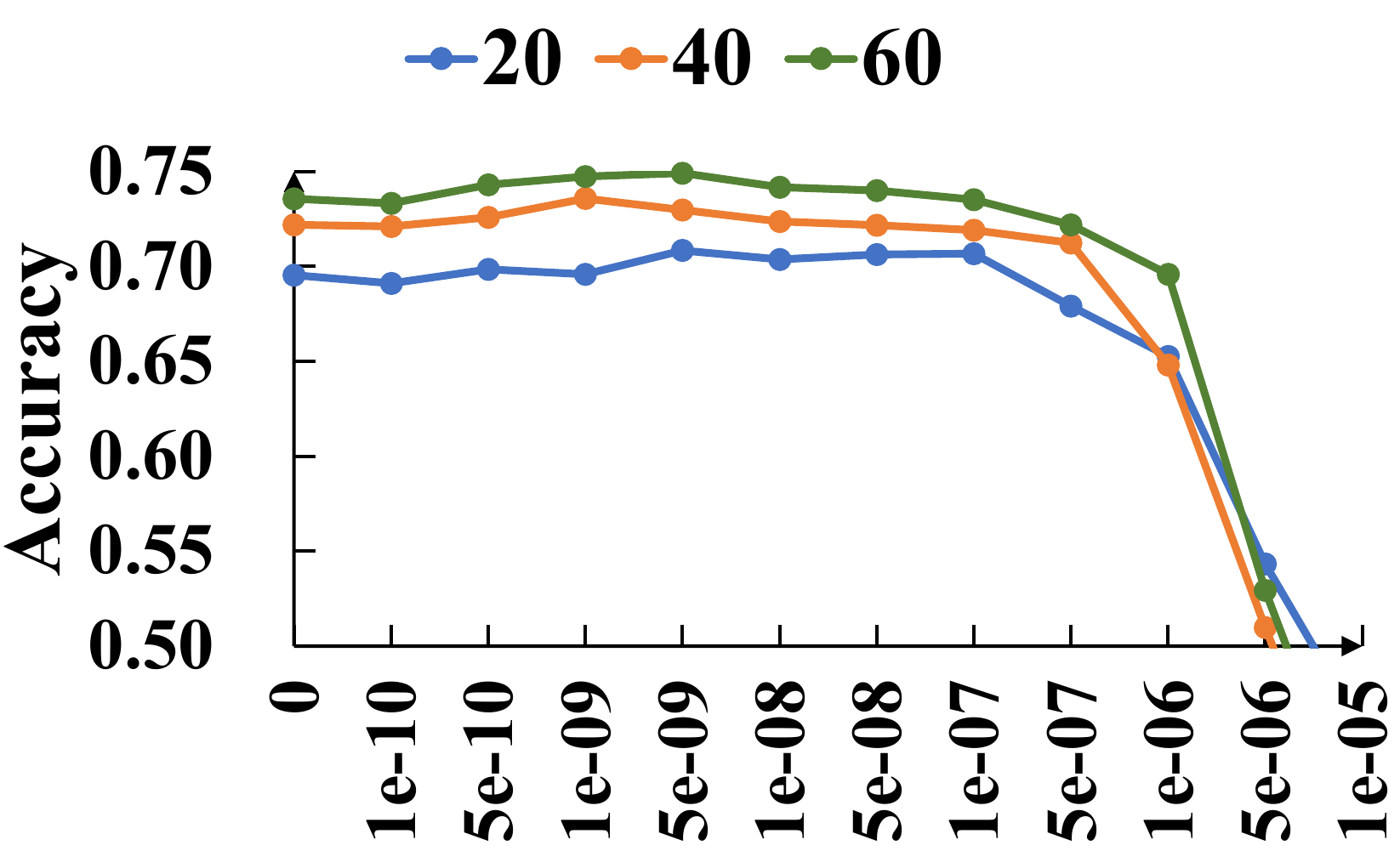}
		%\caption{fig1}
	}
	\subfigure[\textbf{Flickr}]{
		\includegraphics[width=0.22\textwidth]{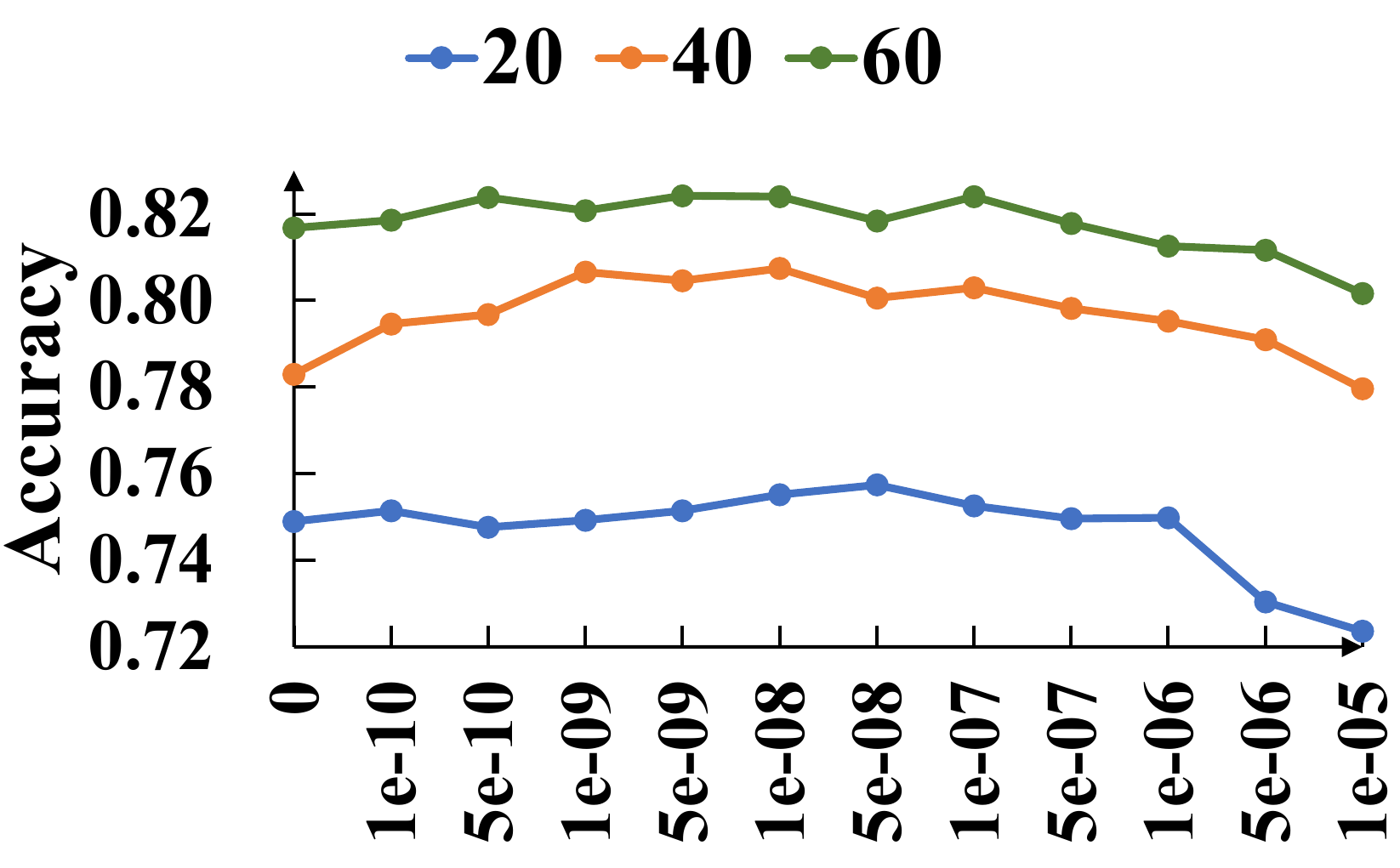}
	}
	\caption{Analysis of parameter \textit{$\beta$}.}
	\vspace{2pt}
	\label{beta_2}
\end{figure}
\begin{figure}[b]
	\centering
	\vspace{2pt}
	\subfigure[\textbf{UAI2010}]{
		\includegraphics[width=0.22\textwidth]{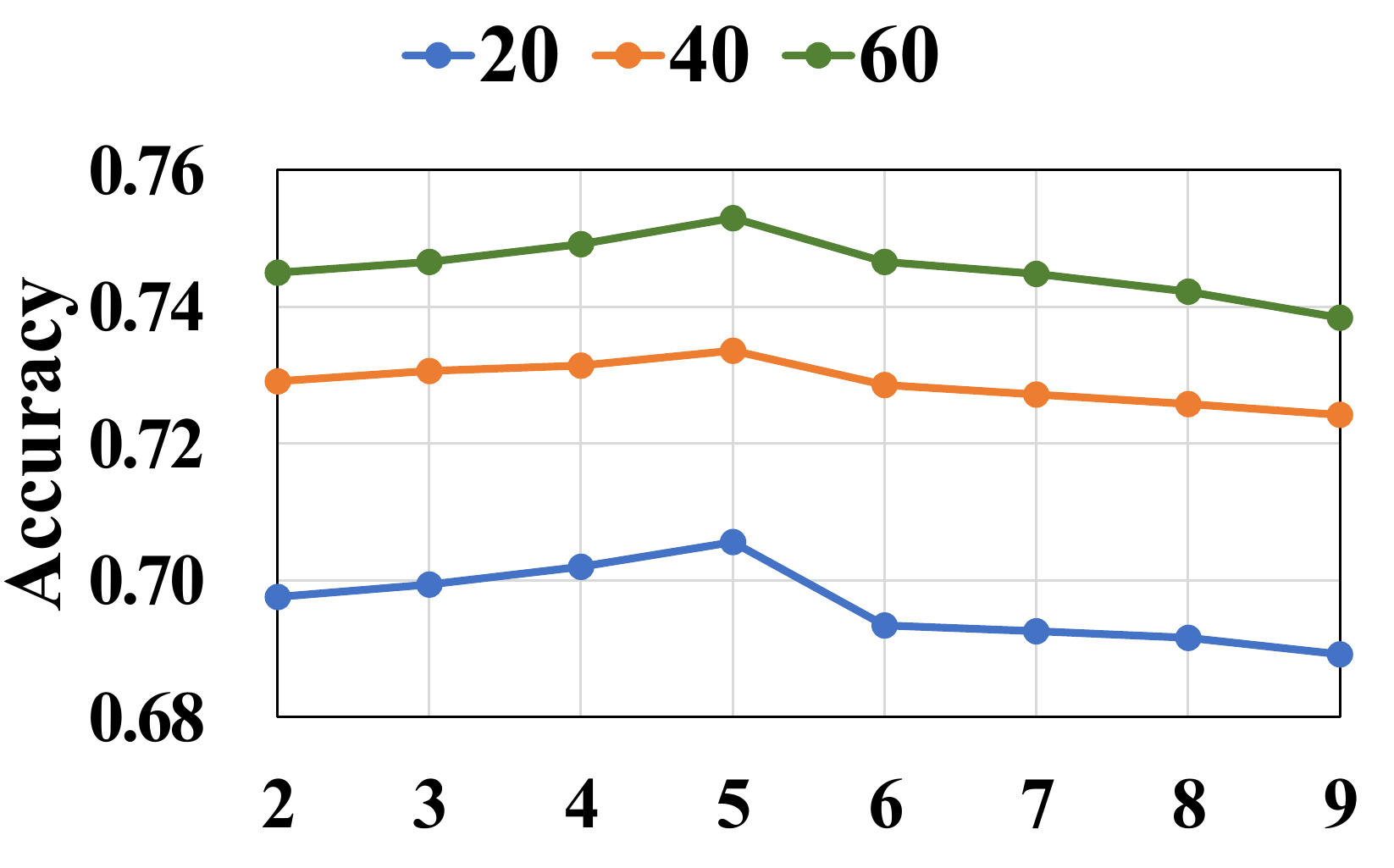}
		%\caption{fig1}
	}
	\subfigure[\textbf{Flickr}]{
		\includegraphics[width=0.22\textwidth]{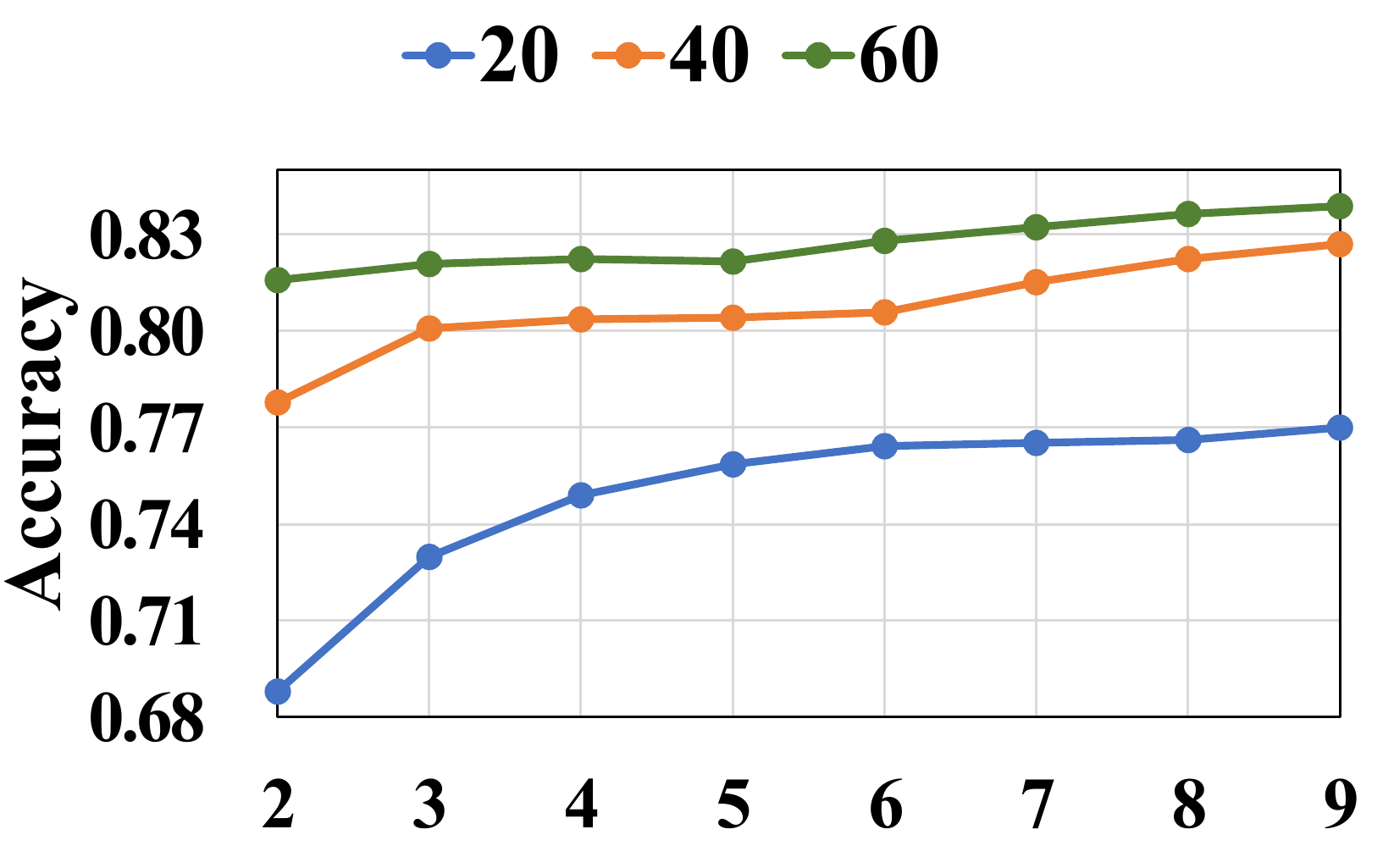}
	}
	\caption{Analysis of parameter \textit{k}.}
	\vspace{2pt}
	\label{k_2}
\end{figure}
\subsubsection{Analysis of attention trends} \label{trends}
Following the setting from the Sectin 4.5, we give the additional analysis of attention trends on
the other four datasets in Figure \ref{attention_change_2}. The changing process of attention values follows the same way with the results in Figure \ref{attention_change}. What's more, the final learned attention values are consist with the corresponding distributions in Figure \ref{boxplot}, from which we can further verify the effectiveness of attention mechanism.

\begin{table*}[!t]
	\caption{Model Hyperparameters.}
	\label{hyperparameters}
	\begin{tabular}{c|c||ccccccccc}
		\hline
		Datasets&L/C&nhid1&nhid2&dropout&lr&weight-decay&epoch$_{max}$&\textit{k}&$\gamma$&$\beta$\\
		\hline
		\multirow{3}{*}{Citeseer}
		&20&768&256&0.5&0.0005&5e-3&25&7&0.001&5e-10\\
		&40&768&128&0.5&0.0005&5e-3&25&7&0.001&5e-8\\
		&60&768&128&0.5&0.0005&5e-3&25&7&0.001&5e-8\\
		\hline
		\multirow{3}{*}{UAI2010}
		&20&512&128&0.5&0.0005&5e-4&50&5&0.001&1e-9\\
		&40&512&128&0.5&0.0005&5e-4&70&5&0.01&1e-9\\
		&60&512&128&0.5&0.0005&1e-5&70&5&0.01&1e-9\\
		\hline
		\multirow{3}{*}{ACM}
		&20&768&256&0.5&0.0005&5e-4&20&5&0.001&1e-8\\
		&40&768&256&0.5&0.0005&5e-4&20&5&0.001&1e-8\\
		&60&768&256&0.5&0.0001&6e-4&30&5&0.001&1e-8\\
		\hline
		\multirow{3}{*}{BlogCatalog}
		&20&512&128&0.5&0.0002&1e-5&55&5&0.001&5e-8\\
		&40&512&128&0.5&0.0005&5e-4&40&5&0.001&5e-8\\
		&60&512&128&0.5&0.0005&8e-4&50&5&0.01&5e-8\\
		\hline
		\multirow{3}{*}{Flickr}
		&20&512&128&0.5&0.0003&5e-4&60&5&0.01&1e-10\\
		&40&512&128&0.5&0.0005&1e-5&40&5&0.01&1e-10\\
		&60&512&128&0.5&0.0005&5e-4&40&5&0.01&1e-10\\
		\hline
		\multirow{3}{*}{CoraFull}
		&20&512&32&0.5&0.001&5e-4&300&6&0.0001&1e-10\\
		&40&512&32&0.5&0.001&5e-4&300&6&0.00001&1e-10\\
		&60&512&32&0.5&0.001&5e-4&300&6&0.0001&1e-10\\
		\hline
	\end{tabular}
\end{table*}

\subsubsection{Parameters Study} \label{ps}
To further check the stability and applicability of parameters \textit{$\gamma$} and \textit{$\beta$}, we show the corresponding results on UAI2010 and Flickr datasets in Figure \ref{alpha_2} and Figure \ref{beta_2}. Combining the results in Section \ref{sec:parameter study}, we can see the consistency and disparity constraints have stable performance on a large range while the performance with disparity constraint may decrease when \textit{$\beta$} is larger than some suitable boundary.
\balance
In Figure \ref{k_2}, we test the impact of \textit{k} in \textit{k}-nearest neighbor graph on UAI2010 and Flicker datasets. For UAI2010, the preformance increases first and then starts to decrease 2 to 10. And for Flickr, a larger \textit{k} may import richer structural information for feature graph which also profits AM-GCN.

\end{document}